\documentclass[10pt,journal,compsoc]{IEEEtran}

\ifCLASSINFOpdf
\else
\usepackage[dvips]{graphicx}
\fi
\usepackage{url}

\usepackage[colorlinks,hypertexnames=false,linkcolor=blue,anchorcolor=blue,
citecolor=blue]{hyperref}
\usepackage{graphicx}
\usepackage{booktabs,tabularx,multirow,threeparttable}
\usepackage{float}
\usepackage{subfigure}
\usepackage{xcolor}
\usepackage{ragged2e}
\usepackage{amsthm,amsmath,amsfonts,amssymb,bm}
\ifCLASSOPTIONcompsoc
  \usepackage[nocompress]{cite}
\else
  \usepackage{cite}
\fi

\newcommand{\hl}[1]{{#1}}
\newcommand{\revii}[1]{{#1}}

\begin{document}
\title{Towards Vision-Language Geo-Foundation Model: A Survey}
\author{Yue~Zhou~\IEEEmembership{Member,~IEEE,} Zhihang~Zhong, Litong~Feng, 
Wayne~Zhang, \\Xue~Jiang~\IEEEmembership{Senior Member,~IEEE,} 
Junchi~Yan~\IEEEmembership{Senior Member,~IEEE,} Xue~Yang*~\IEEEmembership{Member,~IEEE}
    

\thanks{Yue~Zhou is with the School of GeoAI and Hindon STAI Institute, Key Laboratory of Geographic Information Science (Ministry of Education), East China Normal University, Shanghai. Zhihang~Zhong and Junchi~Yan are with the School of AI, Shanghai Jiao Tong University. Litong~Feng, Wayne~Zhang are with SenseTime Research. Xue~Jiang is with Department of Electronic Engineering, Shanghai Jiao Tong University. Xue Yang is with the School of Automation and Intelligent Sensing, Shanghai Jiao Tong University (Email: yangxue-2019-sjtu@sjtu.edu.cn).\\
Corresponding author: Xue Yang.

}

\thanks{This work was supported by the Shanghai Science and Technology Program (25ZR1402268), and the National Natural Science Foundation of China (62506229, 62571315).}
}

\markboth{}%
{Zhou \MakeLowercase{\textit{et al.}}: Towards Vision-Language Geo-Foundation Model: A Survey}

\IEEEtitleabstractindextext{
\justify
\begin{abstract}
Vision-Language Foundation Models (VLFMs) have made remarkable progress on various multimodal tasks, such as image captioning, image-text retrieval, visual question answering, and visual grounding. However, most methods rely on training with general image datasets, and the lack of geospatial data leads to poor performance on earth observation. Numerous geospatial image-text pair datasets and VLFMs fine-tuned on them have been proposed recently. These new approaches aim to leverage large-scale, multimodal geospatial data to build versatile intelligent models with diverse geo-perceptive capabilities, which we refer to as Vision-Language Geo-Foundation Models (VLGFMs). This paper thoroughly reviews VLGFMs, summarizing and analyzing recent developments in the field. In particular, we introduce the background and motivation behind the rise of VLGFMs, highlighting their unique research significance. Then, we systematically summarize the core technologies employed in VLGFMs, including data construction, model architectures, and applications of various multimodal geospatial tasks. Finally, \hl{we synthesize empirical trends, remote-sensing-specific challenges, and future research directions for reliable and measurement-aware VLGFMs. Rather than claiming priority over concurrent surveys, this survey emphasizes an integrated organization of data pipelines, architectures, benchmarked capabilities, and practical limitations.} We keep tracing related works at \url{https://github.com/zytx121/Awesome-VLGFM}.
	
\end{abstract}
	
	}
\maketitle

\begin{figure*}
\centering
\includegraphics[width=18cm]{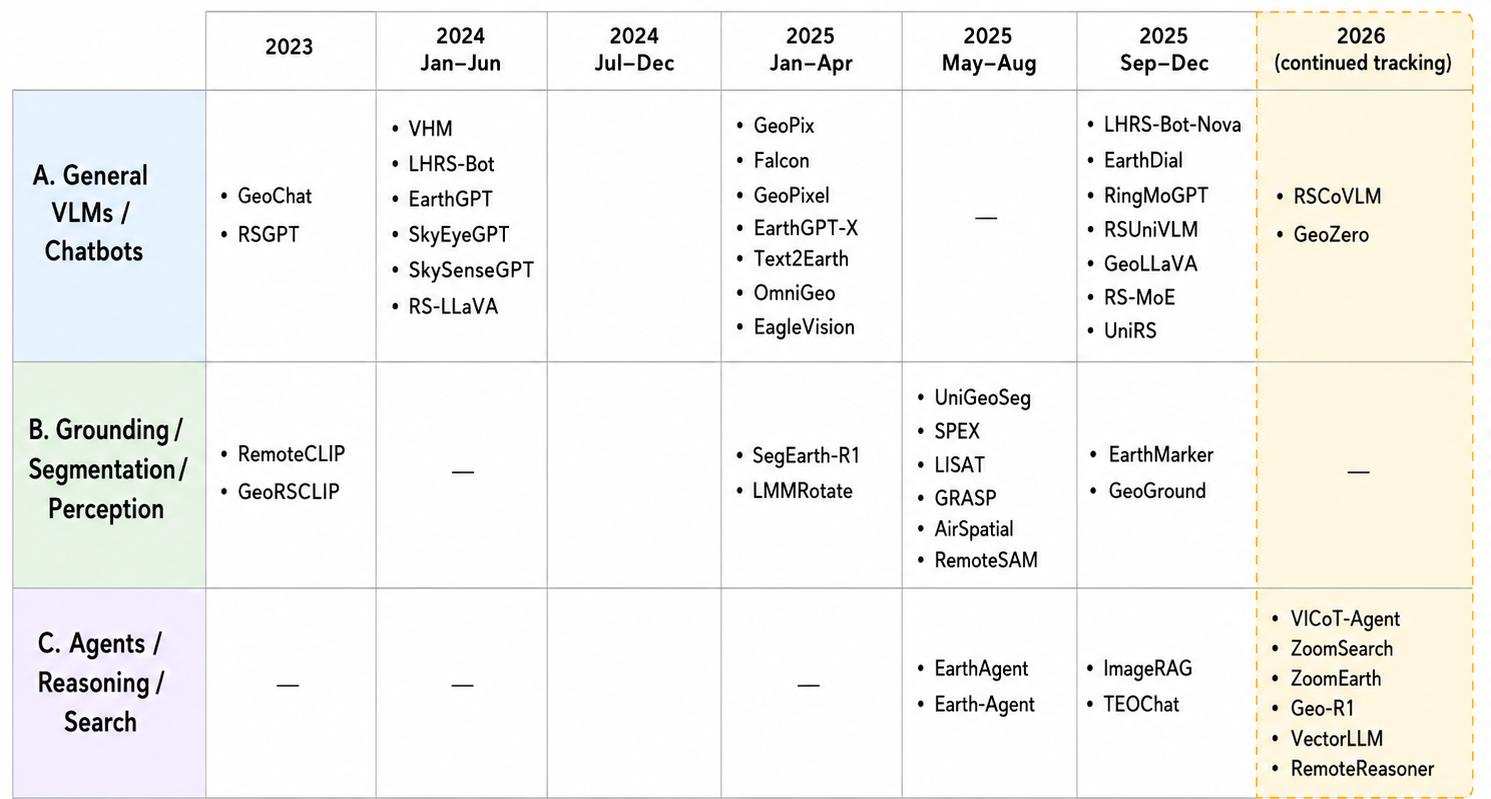}
\caption{\revii{Timeline of representative VLGFMs.} Purple entities indicate projects that currently play a significant role in advancing the development of VLGFMs. For additional resources and daily updates, visit our GitHub page.}
\label{fig:history}
\end{figure*}

\section{Introduction}\label{introduction}

\IEEEPARstart{O}{ver} the past decade, researchers have witnessed significant advancements in nearly all geospatial tasks, such as scene classification \cite{hongdanfeng}, object detection \cite{7560644,9852282}, change detection \cite{8022932}, denoising \cite{denoise}, land use segmentation \cite{8309343}, disaster management \cite{9460988}, and geospatial localization \cite{4587784}, driven by deep learning and other artificial intelligence technologies. However, these models are specifically designed and trained for particular tasks, making them difficult to apply directly to other tasks. \revii{Even for similar tasks, these models often exhibit poor generalization capabilities.}

For example, remote sensing object detection is one of the core tasks for earth observation. It requires manually annotating the position and category of each object, which is a time-consuming and labour-intensive process. Remote sensing images (RSIs) are captured from an overhead perspective by spaceborne or airborne sensors, which present unique viewpoints compared to natural images, leading to the development of oriented object detection. Since this task uses rotated bounding boxes to represent objects, remote sensing datasets annotated with rotated bounding boxes, such as DOTA \cite{Xia_2018_dota}, are required to support its training. Furthermore, the model architecture~\cite{Yang2021R3Det}, loss functions~\cite{Yang2021Rethinking}, post-processing functions, and acceleration operators~\cite{zhou2019iou} must be modified based on standard object detection \cite{mmrotate}. From this perspective, the application scenarios of specific models seem quite limited, lacking the ability to generalize across different tasks, even between two similar tasks.

To reduce the resource waste associated with the training of specific models from scratch for each task, foundational models~\cite{li2024visionlanguage} have emerged. These models are pre-trained on large-scale imagery, allowing them to handle various vision tasks with only fine-tuning small-scale custom datasets. In the field of remote sensing, prior studies on purely visual foundational models reveal immense potential for a generic model for earth observation, which is called Vision Geo-Foundation Models (VGFMs). VGFMs demonstrate remarkable generalization capabilities in a thorough evaluation, from single- to multi-modal, and static to temporal \cite{guo2024skysense}. Although these models exhibit strong perceptual capabilities, they lack the ability to perform reasoning like humans. For instance, without corresponding annotated samples to support VGFM training, it cannot determine the specific function of a building in a remote sensing image by considering the surrounding environment and common sense, whereas humans can. Similarly, without annotated samples,  VGFM cannot identify the brand or model of the car based on its features in aerial imagery, whereas humans can.

Recently, the evolution of Large Language Models (LLMs) has revolutionized human-machine interaction. LLMs like BERT \cite{devlin2019bert} take advantage of large amounts of textual data to develop reasoning skills, showing substantial generalization capabilities on various tasks in natural language processing. However, LLMs process only discrete textual data and cannot handle images, while visual foundation models, though capable of processing image data, lack inferential abilities. The concept of vision-language foundation models (VLFMs) has been introduced to bridge these gaps. These innovative models are designed to perform perception and reasoning, integrating inputs from text and images. \hl{Since the release of GPT-4 Vision, research on VLFMs has accelerated rapidly.} Substantial efforts have been made in VLFMs, which are typically divided into contrastive, conversational, and generative paradigms. Below, we will briefly introduce the most influential works in three directions. Specifically, CLIP \cite{pmlr-v139-radford21a} adopts the contrastive paradigm, projecting visual and textual information into a unified representational space, thereby facilitating a bridge for downstream vision-language tasks. LLaVA \cite{llava} embodies the conversational paradigm, endowing LLMs with the capability to recognize context in textual and visual modalities. \hl{Stable Diffusion} \cite{rombach2022highresolution} \hl{is a representative generative model that produces high-quality images from textual descriptions and has influenced text-conditioned visual synthesis.}

When VLFMs are applied to earth observation, this paper refers to them as Vision-Language Geo-Foundation Models (VLGFMs). To date, VLGFMs can also be categorized into contrastive, conversational, and generative types. \revii{Figure~\ref{fig:history} presents representative VLGFMs and their release dates.} VLGFMs first appeared in the second quarter of 2023, and the literature has since grown rapidly. Notably, current innovations in VLGFM are primarily focused on collecting training data, with relatively few modifications to the model architecture. Most efforts involve fine-tuning based on the LLaVA \cite{llava} and MiniGPT-4 \cite{zhu2023minigpt4} frameworks using custom-built remote sensing instruction-following sets.

As VLGFM continues to advance rapidly and demonstrate impressive results, tracking and comparing recent research on VLGFM is worthwhile. It transforms conventional human-computer interaction, which relies on predefined program interfaces, by enabling end-to-end interaction with humans through natural language dialogue. \hl{Several recent surveys have discussed remote sensing vision-language models, multimodal large language models, and remote sensing foundation models from complementary perspectives. Therefore, this paper does not frame its contribution as the first survey. Instead, it positions itself as a capability-oriented synthesis of VLGFMs, with a focus on how data construction, model architecture, benchmark design, and geo-specific capabilities jointly shape model behavior.}

\noindent \textbf{Contribution.}
In light of the rapid advancements and promising results of VLGFM, we have composed this survey to acquaint researchers with the fundamental concepts, principal methodologies, and current progress of VLGFMs. \hl{Compared with existing surveys, as summarized in Table~\ref{tab:survey-comparison}, this work emphasizes four aspects: a systematic comparison of data construction strategies, a unified review of contrastive, conversational, and generative architectures, a hierarchical capability taxonomy tailored to earth observation, and a benchmark-oriented discussion of empirical trends and limitations.} It also provides a comparative analysis of the background and related concepts, including VGFMs and remote sensing LLM-powered agents. Furthermore, we categorize the demonstrated capabilities of VLGFMs into three levels. \hl{We hope this organization helps readers understand not only which models have been proposed, but also why different data and architectural choices lead to different capabilities.}

\begin{table*}[!t]
\centering
\scriptsize
\caption{\hl{Comparison with representative surveys related to remote sensing vision-language and multimodal foundation models. A check mark denotes substantive coverage.}}
\label{tab:survey-comparison}
\resizebox{\linewidth}{!}{%
\begin{tabular}{p{3.5cm}|c|c|c|c|c|c|p{2.0cm}|p{3.7cm}}
\hline
\textbf{Survey} & \textbf{Data} & \textbf{Arch.} & \textbf{Capabilities} & \textbf{Benchmarks} & \textbf{Task results} & \textbf{Temporal} & \textbf{Coverage / corpus} & \textbf{Primary emphasis} \\
\hline
Li et al.~\cite{li2024visionlanguage} & \checkmark & \checkmark &  & \checkmark &  &  & \hl{2023} & \hl{Early progress and trends in remote sensing VLMs} \\
Weng et al.~\cite{weng2025vlmrs} & \checkmark & \checkmark & \checkmark & \checkmark &  &  & \hl{May 2025} & \hl{Contrastive learning, instruction tuning, and text-conditioned generation} \\
Tao et al.~\cite{tao2025advancements} & \checkmark & \checkmark & \checkmark & \checkmark &  &  & \hl{2025} & \hl{Datasets, capabilities, and enhancement techniques} \\
Liu et al.~\cite{liu2024rstvlmsurvey} & \checkmark & \checkmark & \checkmark & \checkmark &  & \checkmark & \hl{Dec. 2024} & \hl{Temporal and spatio-temporal vision-language models} \\
Huang et al.~\cite{huang2025rsfmsurvey} & \checkmark & \checkmark &  & \checkmark &  & \checkmark & \hl{Mar. 2025} & \hl{Remote sensing foundation models from vision to multimodality} \\
Xu et al.~\cite{xu2026eomllmsurvey} & \checkmark & \checkmark & \checkmark & \checkmark &  &  & \hl{July 2025 online} & \hl{Frameworks and core technologies of Earth-observation MLLMs} \\
\textbf{This survey} & \revii{\checkmark} & \revii{\checkmark} & \revii{\checkmark} & \revii{\checkmark} & \revii{\checkmark} & \revii{\checkmark} & \revii{Updated Jun. 2026; 128 unique cited records} & \hl{Capability-oriented synthesis with quantitative and photogrammetric analysis} \\
\hline
\end{tabular}}
\end{table*}

\noindent \textbf{Survey methodology.}
\hl{We searched IEEE Xplore, Web of Science, Scopus, Google Scholar, and arXiv, complemented by the reference lists of relevant surveys and major remote sensing, computer vision, and machine-learning venues. Queries combined ``remote sensing'' or ``earth observation'' with ``vision-language model'', ``multimodal large language model'', ``geo-foundation model'', ``visual grounding'', ``visual question answering'', ``image-text retrieval'', ``text-to-image generation'', and ``remote sensing agent''.} \revii{The main search period begins in January 2023, when VLGFMs started to emerge, and the literature collection was updated through June 2026 during the second-round revision. Earlier seminal work was retained for definitions, datasets, metrics, standards, and technical context. The working bibliography contained 319 candidates; after relevance screening, full-text review, and a final duplicate-reference audit, the manuscript cites 128 unique records.} \hl{We included work that introduces or systematically evaluates a reusable remote-sensing-oriented vision-language foundation model, dataset, benchmark, or agent. We excluded purely visual foundation models, conventional single-task networks without a reusable vision-language representation, and papers lacking a clear geospatial vision-language contribution. Works selected for detailed narrative discussion additionally had to introduce a new model or dataset, cover a distinct capability, establish a frequently used benchmark, or materially change the state of the art; other eligible records remain represented in the tables and repository.}

\noindent \textbf{Survey pipeline.}
In Section 2, we provide background knowledge, including definitions, datasets, metrics, and related research areas. In Section 3, we conduct a comprehensive review of various methods based on different data collection approaches, network architectures, and capabilities. In Section 4, we identify challenges and future directions.

\begin{table*}[!h]
\centering 
\vskip 1mm
\caption{\hl{Model architecture of VLGFMs.} Sort models by date of initial release. Abbreviations IS, IC, VC, IR, OC, REC, REG, RES, and GM in the table stand for image scene, image captioning, video captioning, image retrieval, object counting, referring expression comprehension, referring expression generation, referring expression segmentation, and geometric measurement, respectively. For each model, we indicate the LLM used in its best configuration as shown in the original paper ($\lozenge$: training from scratch; $\blacklozenge$: fine-tuning; $\blacktriangle$: fine-tuning with parameter-efficient fine-tuning techniques \cite{si2024flora}; $\bigstar$: frozen). }
\vskip 1mm
\resizebox{\linewidth}{!}{
\begin{tabular}{c|cccccccc}
\hline
Model &LLM & Visual Encoder & Connector & Capabilities  & Hardware\\ 
\hline
RemoteCLIP & \multirow{2}{*}{-} & \multirow{2}{*}{ViT-L (CLIP)$^\blacklozenge$}  & \multirow{2}{*}{-} & IS, IR, & \multirow{2}{*}{4*3090Ti (24G)}\\ 
\cite{liu2024remoteclip}&  &  &  & OC &  \\
\hline
GeoRSCLIP & \multirow{2}{*}{-} & \multirow{2}{*}{ViT-H (CLIP)$^\blacklozenge$} & \multirow{2}{*}{-} & \multirow{2}{*}{IS, IR} & 1*A100 (40G) +\\ 
\cite{zhang2024rs5m}&  &  &  & & 1*A100 (80G) \\
 \hline
RSGPT & \multirow{2}{*}{Vicuna-13B$^\bigstar$} & \multirow{2}{*}{ViT-G (EVA)$^\bigstar$} & a Q-Former$^\blacklozenge$ + & \multirow{2}{*}{VQA, IC} & \multirow{2}{*}{8*A100 (80G)}\\ 
\cite{hu2023rsgpt}&  &  & a linear layer$^\blacklozenge$ & &  \\
 \hline
GeoChat & \multirow{2}{*}{Vicuna-v1.5-7B$^\blacktriangle$} & \multirow{2}{*}{ViT-L (CLIP)$^\bigstar$} & \multirow{2}{*}{a two-layer MLP$^\bigstar$}   & IS, VQA, IC, & \multirow{2}{*}{3*A100 (40G)}\\ 
\cite{kuckreja2023geochat}& &  &  & REC, REG &  \\
\hline
GRAFT & \multirow{2}{*}{-} & \multirow{2}{*}{ViT-B (CLIP)$^\blacklozenge$} & \multirow{2}{*}{-} & IS, IR, VQA, & \multirow{2}{*}{Unknown} \\ 
\cite{mall2023remote}&  &  &  &  RES &  \\
\hline
SkyScript & \multirow{2}{*}{-} & \multirow{2}{*}{ViT-L (CLIP)$^\blacklozenge$} & \multirow{2}{*}{-} & \multirow{2}{*}{IS, IR} & \multirow{2}{*}{4*A100 (80G)} \\ 
\cite{wang2024skyscript}& &  &  &   &  \\
\hline
\multirow{2}{*}{SkyEyeGPT}& \multirow{3}{*}{LLaMA2-7B$^\blacktriangle$} & \multirow{3}{*}{ViT-G (EVA-CLIP)$^\bigstar$} & \multirow{3}{*}{a linear layer$^\blacktriangle$} & IS, VQA, IC,  & \multirow{3}{*}{4*3090Ti (24G)}\\
\multirow{2}{*}{\cite{zhan2024skyeyegpt}}&  &  &  &  REC, REG, &  \\
& &  &   & VC &  \\
\hline
EarthGPT& \multirow{2}{*}{LLaMA2-7B$^\blacklozenge$} & ViT-L (DINOv2)$^\bigstar$ +  & \multirow{2}{*}{a linear layer$^\blacklozenge$} & IS, VQA, IC,  & \multirow{2}{*}{16*A100 (80G)}\\
\cite{zhang2024earthgpt}&  &  ConvNeXt-L (CLIP)$^\bigstar$ & & REC, REG  \\
 \hline
LHRS-Bot& \multirow{2}{*}{LLaMA2-7B$^\blacktriangle$} & \multirow{2}{*}{ViT-L (CLIP)$^\bigstar$} & \multirow{2}{*}{a Vision Perceiver$^\blacklozenge$} & IS, VQA, IC, & \multirow{2}{*}{8*V100 (32G)}\\
\cite{muhtar2024lhrsbot}&  &  &  & REC &  \\
\hline
RS-CapRet& \multirow{2}{*}{LLaMA2-7B$^\bigstar$} & \multirow{2}{*}{ViT-L (CLIP)$^\blacklozenge$} & \multirow{2}{*}{three linear layers $^\blacklozenge$} & \multirow{2}{*}{IR, IC} & \multirow{2}{*}{Unknown}\\
\cite{silva2024large}&  &  &  &  &  \\
\hline
\multirow{2}{*}{H$^2$RSVLM} & \multirow{3}{*}{Vicuna-v1.5-7B$^\blacklozenge$}  & \multirow{3}{*}{ViT-L (CLIP)$^\blacklozenge$}  & \multirow{3}{*}{a two-layer MLP$^\blacklozenge$}  & IS, IM, VQA, & \multirow{3}{*}{16*A100 (80G)}\\
\multirow{2}{*}{\cite{pang2024h2rsvlm}} &  &  &  &  IC, REC, REG, &  \\
&  &  &  & OC, GM, SP &  \\
\hline
RS-LLaVA & \multirow{2}{*}{Vicuna-v1.5-13B$^\blacktriangle$}  & \multirow{2}{*}{ViT-L (CLIP)$^\blacktriangle$}  & \multirow{2}{*}{a two-layer MLP$^\blacktriangle$}  &  \multirow{2}{*}{VQA, IC} & \multirow{2}{*}{2*A6000 (48G)}\\
\cite{rsllava} &  &  &  &   &  \\
\hline
SkySenseGPT & \multirow{2}{*}{Vicuna-v1.5-7B$^\blacktriangle$}  & \multirow{2}{*}{ViT-L (CLIP)$^\bigstar$}  & \multirow{2}{*}{a two-layer MLP$^\bigstar$}  &  VQA, IC, & \multirow{2}{*}{4*A100 (40G)}\\
\cite{luo2024skysensegpt} &  &  &  & OVD, REC, REG &  \\
\hline
RSUniVLM & \multirow{2}{*}{Qwen2-0.5B$^\blacklozenge$}  & \multirow{2}{*}{SigLIP-400M$^\bigstar$}  & \multirow{2}{*}{a two-layer MLP$^\bigstar$}  & VQA, RES, & \multirow{2}{*}{4*A40 (40G)}\\
\cite{liu2024rsunivlm} &  &  &  & IS, IC, REC  &  \\
\hline
EarthDial & \multirow{2}{*}{Phi-3-mini$^\blacklozenge$}  & \multirow{2}{*}{InternViT-300M$^\blacklozenge$}  & \multirow{2}{*}{a two-layer MLP$^\blacklozenge$}  & VQA, OVD, & \multirow{2}{*}{8*A100 (80G)}\\
\cite{soni2024earthdial} &  &  &  & IS, IC, REC  &  \\
\hline

RingMo-Agent & \multirow{2}{*}{DeepSeekMoE-3B$^\blacktriangle$}  & \multirow{2}{*}{SigLIP-SO400M-384$^\bigstar$}  & \multirow{2}{*}{a two-layer MLP$^\bigstar$}  & VQA, OVD, & \multirow{2}{*}{8*A100 (80G)}\\
\cite{hu2025ringmoagent} &  &  &  & IS, IC  &  \\
\hline
RSCoVLM & \multirow{2}{*}{Qwen2.5-7B$^\blacklozenge$}  & \multirow{2}{*}{Modified ViT$^\bigstar$}  & \multirow{2}{*}{a two-layer MLP$^\bigstar$}  & VQA, OVD, & \multirow{2}{*}{8*A800 (80G)}\\
\cite{li2025rscovlm} &  &  &  & IS, REC  &  \\
\hline
GeoZero & \multirow{2}{*}{Qwen3-8B$^\blacktriangle$}  & \multirow{2}{*}{SigLIP2-SO-400M$^\bigstar$}  & \multirow{2}{*}{a two-layer MLP$^\bigstar$}  &  VQA, IC, & \multirow{2}{*}{8*A100 (40G)}\\
\cite{wang2025geozero} &  &  &  & IS, REC, VQA  &  \\
\hline
\end{tabular}}
\label{Table0}
\end{table*}

\begin{table*}[!t]
\centering 
\vskip 1mm
\caption{\revii{Evaluation datasets used by VLGFMs. IC, VQA, VG, IS, and IR denote image captioning, visual question answering, visual grounding, image scene classification, and image retrieval, respectively. A dataset used by more than one model is shown in \textcolor{blue}{blue}; one used by more than two models is shown in \textcolor{red}{red}.}}
\vskip 1mm
\resizebox{\linewidth}{!}{
\begin{tabular}{c|cccccc}
\hline
Model & IC & VQA & VG & IS  & IR & Others \\ 
\hline
 \multirow{5}{*}{RemoteCLIP}  &  \multirow{6}{*}{-} & \multirow{6}{*}{-} & \multirow{6}{*}{-} & \textcolor{blue}{PatternNet}, \textcolor{red}{EuroSAT}, &   & \multirow{6}{*}{RemoteCount}\\ 
 \multirow{5}{*}{\cite{liu2024remoteclip}}  &  & &  & OPTIMAL-31, RSC11, &  \textcolor{red}{RSITMD},& \\
 &  & &  & RSICB128, MLRSNet, & \textcolor{red}{RSICD}, & \\
  &  & &  & \textcolor{blue}{RSI-CB256}, \textcolor{red}{RESISC45}, & \textcolor{red}{UCM} & \\
  \multirow{5}{*}  &  & &  & WHU-earth, RS2800, &  & \\
    &  & &  & \textcolor{red}{WHU-RS19} &  & \\
\hline
GeoRSCLIP & \multirow{2}{*}{-} & \multirow{2}{*}{-} & \multirow{2}{*}{-} & \textcolor{red}{EuroSAT}, \textcolor{red}{AID},  & \textcolor{red}{RSITMD}, & \multirow{2}{*}{-} \\ 
\cite{liu2024remoteclip} &  & &  & 	\textcolor{red}{RESISC45} &  \textcolor{red}{RSICD} & \\
\hline
RSGPT & \textcolor{red}{UCM-Captions} & \textcolor{red}{RSVQA-LR} & \multirow{2}{*}{-} & \multirow{2}{*}{-} & \multirow{2}{*}{-} & \multirow{2}{*}{RSIEval} \\ 
\cite{hu2023rsgpt} & \textcolor{red}{Sydney-Captions} & \textcolor{red}{RSVQA-HR} &  & & & \\
\hline
GeoChat & \multirow{2}{*}{-} & \textcolor{red}{RSVQA-LR} & \multirow{2}{*}{-} & \textcolor{red}{AID}, & \multirow{2}{*}{-} & \multirow{2}{*}{GeoChat-Bench}\\ 
\cite{kuckreja2023geochat} &  & \textcolor{red}{RSVQA-HR} & & UCM & &  \\
\hline
GRAFT & \multirow{2}{*}{-} & \multirow{2}{*}{\textcolor{red}{RSVQA-HR}} & \multirow{2}{*}{-} & \textcolor{red}{EuroSAT}, BEN,  & EuroSAT, & \multirow{2}{*}{NAIP-OSM}\\ 
\cite{mall2023remote} &  &  & & SAT-4, SAT-6 & BEN & \\
\hline
\multirow{2}{*}{SkyCLIP} & \multirow{3}{*}{-} & \multirow{3}{*}{-} & \multirow{3}{*}{-} & \textcolor{blue}{PatternNet}, \textcolor{red}{EuroSAT}, & \textcolor{red}{RSITMD}, & \multirow{3}{*}{SkyScript}\\ 
\multirow{2}{*}{\cite{wang2024skyscript}} &  & &  & \textcolor{red}{AID}, \textcolor{blue}{fMoW}, \textcolor{red}{RESISC45}, & \textcolor{red}{RSICD},& \\
 &  & &  & MillionAID, \textcolor{blue}{RSI-CB256}  & \textcolor{red}{UCM} & \\
\hline
\multirow{2}{*}{SkyEyeGPT} & \textcolor{red}{UCM-Captions},   & \multirow{2}{*}{\textcolor{red}{RSVQA-LR},} & \multirow{2}{*}{\textcolor{blue}{RSVG},} &\multirow{3}{*}{-} & \multirow{3}{*}{-}&\multirow{3}{*}{-} \\ 
\multirow{2}{*}{\cite{zhan2024skyeyegpt}} & \textcolor{red}{Sydney-Captions}, & \multirow{2}{*}{\textcolor{red}{RSVQA-HR}} & \multirow{2}{*}{\textcolor{red}{DIOR-RSVG}} & & & \\
 & \textcolor{blue}{RSICD}, CapERA &  & & &  & \\
\hline
EarthGPT & \multirow{2}{*}{\textcolor{blue}{NWPU-Captions}} & CRSVQA & \textcolor{red}{DIOR-RSVG} & \textcolor{red}{RESISC45}  & \multirow{2}{*}{-}& \multirow{2}{*}{-}\\ 
\cite{zhang2024earthgpt} &  & \textcolor{red}{RSVQA-HR} & MAR20 &CLRS, NaSC-TG2 & & \\
\hline
\multirow{3}{*}{LHRS-Bot} & \multirow{4}{*}{-} & \multirow{3}{*}{\textcolor{red}{RSVQA-LR},} & \multirow{3}{*}{\textcolor{blue}{RSVG},} & \textcolor{red}{AID}, \textcolor{red}{WHU-RS19},   & \multirow{4}{*}{-}& \multirow{4}{*}{-}\\ 
\multirow{3}{*}{\cite{muhtar2024lhrsbot}} &  & \multirow{3}{*}{\textcolor{red}{RSVQA-HR}} & \multirow{3}{*}{\textcolor{red}{DIOR-RSVG}} & \textcolor{red}{SIRI-WHU}, \textcolor{red}{EuroSAT}, &  & \\
 &  &  & & \textcolor{red}{RESISC45}, \textcolor{blue}{fMoW}, &  & \\
  &  &  & & \textcolor{blue}{METER-ML} & & \\
\hline
\multirow{3}{*}{RS-CapRet} & \textcolor{red}{UCM-Captions},  & \multirow{4}{*}{-} & \multirow{4}{*}{-} & \multirow{4}{*}{-} & \multirow{3}{*}{\textcolor{red}{RSICD}} & \multirow{4}{*}{-} \\ 
\multirow{3}{*}{\cite{silva2024large}} & \textcolor{red}{Sydney-Captions}, &  & & & \multirow{3}{*}{\textcolor{red}{UCM}} & \\
 & \textcolor{blue}{NWPU-Captions}, &  & & &   & \\
  & \textcolor{blue}{RSICD} &  & & &   & \\
\hline
\multirow{2}{*}{H$^2$RSVLM} & \multirow{3}{*}{-} & \multirow{2}{*}{\textcolor{red}{RSVQA-LR}} & \multirow{3}{*}{\textcolor{red}{DIOR-RSVG}} & NWPU, \textcolor{blue}{METER-ML}, &  \multirow{3}{*}{-} & \multirow{2}{*}{RSSA}\\
\multirow{2}{*}{\cite{pang2024h2rsvlm}} &  & \multirow{2}{*}{\textcolor{red}{RSVQA-HR}} & & \textcolor{red}{SIRI-WHU}, \textcolor{red}{AID},  &  &  \multirow{2}{*}{HqDC-Instruct}\\
 &  &  & & \textcolor{red}{WHU-RS19} &  & \\
\hline
RS-LLaVA & \textcolor{red}{UCM-Captions}, & \textcolor{red}{RSVQA-LR}, & \multirow{2}{*}{-} & \multirow{2}{*}{-} & \multirow{2}{*}{-} & \multirow{2}{*}{-}\\
\cite{rsllava} & UAV & RSIVQA &  &  &   &\\
\hline
SkySenseGPT & &\textcolor{red}{RSVQA-LR}&&\textcolor{red}{AID}, \textcolor{red}{WHU-RS19},&&FIT-RSFG\\
\cite{luo2024skysensegpt} &  & \textcolor{red}{RSVQA-HR} &  &\textcolor{red}{SIRI-WHU} & & FIT-RSRC\\
\hline
\multirow{2}{*}{RSUniVLM} & \multirow{3}{*}{LEVIR-MCI} & \multirow{2}{*}{\textcolor{red}{RSVQA-LR}} & \multirow{2}{*}{\textcolor{red}{DIOR-RSVG}} & \textcolor{red}{RESISC45}, &  \multirow{3}{*}{-} & WHU-CD,\\
\multirow{2}{*}{\cite{liu2024rsunivlm}} &  & \multirow{2}{*}{\textcolor{red}{RSVQA-HR}} &  \multirow{2}{*}{VRSBench} & \textcolor{red}{SIRI-WHU}, \textcolor{red}{AID},  &  &  Vaihingen, \\
 &  &  & & \textcolor{red}{WHU-RS19} &  & UDD5, VDD \\
\hline
 & &\textcolor{red}{RSVQA-LR},&\textcolor{red}{RSVG}, GeoChat,&\textcolor{red}{AID}, \textcolor{red}{WHU-RS19},&& \\
RSCoVLM &  & \textcolor{red}{RSVQA-HR}, & \textcolor{red}{DIOR-RSVG} &\textcolor{red}{METER-ML}, & & DOTA-v1.0\\
\cite{li2025rscovlm} &  & LRS-VQA, & \textcolor{blue}{VRSBench} &\textcolor{red}{RESISC45}, & & \\
 &  & VRSBench & AVVG & \textcolor{red}{UCM} & & \\ 
\hline
 & \textcolor{red}{UCM-Captions}&&\textcolor{red}{RSVG},&\textcolor{red}{AID}, \textcolor{red}{WHU-RS19},&& \\
GeoZero & \textcolor{red}{Sydney-Captions} & \textcolor{red}{RSVQA-HR}, & \textcolor{red}{DIOR-RSVG} &\textcolor{red}{SIRI-WHU}, & & CHOICE\\
\cite{wang2025geozero} & \textcolor{red}{NWPU-Captions} & XLRSBench & \textcolor{blue}{VRSBench} &\textcolor{red}{RESISC45}, & & \\
 & \textcolor{red}{RSICD} &  &  & \textcolor{red}{EuroSAT},\textcolor{red}{UCM} & & \\ 

\hline
\end{tabular}}
\label{Table1}
\end{table*}

\section{Background}

\textbf{Overview.} In this section, we first introduce the conceptual definition of VLGFM and compare it with related concepts. Then, we provide a historical overview of VLGFM and highlight several representative models. Next, we present the commonly used benchmark datasets and metrics in the VLGFM field. Finally, we review related research domains.

\subsection{Concept Definition}

\textbf{Foundation models}  is a kind of model with broad applicability and general capabilities after training on large-scale data. These models typically have a large number of parameters and strong learning abilities, enabling them to excel in a variety of tasks. Although foundation models may not be ubiquitous at present, they appear poised to become the basis of widespread technological innovations and exhibit the key hallmarks of a general-purpose technology \cite{bommasani2022opportunities}. The field surveyed in this paper can handle multiple tasks simultaneously, and vision-language models that support only a single task are outside the scope of our research. Please refer to \cite{li2024visionlanguage} for those interested in task-specific vision-language models.

\noindent \textbf{Geo-foundation models} are a class of models specifically designed for processing geospatial data through visual information. These models utilize various types of geospatial visual data, such as remote sensing imagery, satellite photos, and aerial images, to perform detailed analysis and support a range of geographic applications. Although this research is currently a popular topic, it is not the focus of this paper. For those interested in purely visual geo-foundation models, please refer to \cite{10282966} for further information.

\noindent \textbf{Vision-language geo-foundation models} are a specialized subset of artificial intelligence models designed for processing and analyzing geospatial data by integrating visual and linguistic information. These models can handle diverse geospatial data sources, such as remote sensing imagery, geographic information system data, and geo-tagged text, leveraging their cross-modal processing capabilities to understand and integrate different types of geospatial information. By combining visual and linguistic modalities, VLGFMs can perform more comprehensive and accurate analyses of geospatial data, making them highly valuable for complex earth observation tasks. This paper focuses on these VLGFMs, excluding those limited to the visual modality or specific tasks. \hl{We further distinguish VLGFMs from remote sensing multimodal large language models (MLLMs). VLGFMs emphasize broad vision-language representation and task transfer for geospatial data, and they may output embeddings, class labels, captions, boxes, masks, or other structured predictions. By contrast, remote sensing MLLMs are usually built around a large language model and emphasize instruction following, open-ended dialogue, tool use, and natural-language reasoning. The two concepts overlap when an MLLM is trained or adapted as a general-purpose geospatial vision-language foundation model, but they should not be treated as identical.}

\subsection{History and Roadmap}

\revii{Before introducing the VLGFM details, it is necessary to review its progress. Figure~\ref{fig:history} summarizes the timeline and lists representative works.} The development of VLFMs in the remote sensing domain started relatively late. It was not until the emergence of the works LLaVA \cite{llava} and MiniGPT-4 \cite{zhu2023minigpt4} in 2023 that VLGFM-related work began to take shape. Most existing conversational VLGFMs have been implemented based on the open-source frameworks released by these two works. Since June 2023, the first batch of VLGFMs began to emerge. Among them, RemoteCLIP \cite{liu2024remoteclip} was the first contrastive VLGFM, supporting image scene classification and image-text retrieval tasks. RSGPT \cite{hu2023rsgpt} was the first conversational VLGFM, supporting image captioning and visual question answering tasks. RS5M \cite{zhang2024rs5m} was the first million-scale remote sensing image-text pair dataset made publicly available. Since October 2023, the number of VLGFM works has steadily increased. Among them, DiffusionSat \cite{khanna2023diffusionsat} is the first high-quality text-to-image generation work in the remote sensing field. It can control the generation of RSIs using conditions such as latitude and longitude, season, cloud cover, and ground sampling distance (GSD). GeoChat \cite{kuckreja2023geochat} is the first conversational VLGFM to support visual grounding task, while SkyEyeGPT \cite{zhan2024skyeyegpt} is the first to support remote sensing video captioning task. H$^2$RSVLM \cite{pang2024h2rsvlm} is the first to introduce the concept of trustworthiness, where the model refuses to answer uncertain questions. SkySenseGPT \cite{luo2024skysensegpt} introduces the concept of graphs to enhance VLGFM's ability to perceive and reason relationships between remote sensing targets as well as its capability for complex comprehension. It is worth mentioning that research on VLGFMs is more akin to a data-centric project rather than a model-centric one. As seen from Table \ref{Table0}, the structure of each model is quite similar, with no significant modifications. However, Table \ref{Table1} shows that the data selection for different VLGFMs varies considerably.

\subsection{Tasks, Datasets, and Metrics}

\noindent$\bullet$
\textbf{Tasks.} VLGFMs are capable of addressing a wide range of earth observation tasks, including image scene classification (IS) \cite{roberts2023satin}, image retrieval (IR) \cite{rsitmd}, image captioning (IC) \cite{sydneycaption}, visual question answering (VQA) \cite{rsvqa}, and visual grounding (VG) \cite{rsvg}. Among these tasks, the VQA task is particularly special as it serves as a means to evaluate most multi-modal tasks in VLGFMs. Specifically, the inputs of VQA include images and text prompts, and the output is text. By using different text prompts, answers for various tasks can be obtained. VG is a more practical task in real-world applications and is mainly divided into two sub-tasks: referring expression comprehension and referring expression generation. Detailed introduction to the various capabilities of VLGFMs will be provided in Section \ref{sec:capabilities}.

\noindent$\bullet$
\textbf{Datasets.} In Table \ref{Table1}, we list the datasets commonly used in experimental validation of existing VLGFMs and categorize them into three types based on usage frequency: recommended, common, and rare. Red, blue, and black fonts represent these categories. For the image captioning task, the recommended datasets are UCM-Captions and Sydney-Captions \cite{sydneycaption}; the commonly used datasets are RSICD \cite{rsicd} and NWPU-Captions \cite{nwpucaption}; and the rare datasets are CapERA \cite{capera} and UAV \cite{uavcaption}. For the VQA task, the recommended datasets are RSVQA-LR and RSVQA-HR \cite{rsvqa}. The recommended dataset for the visual grounding task is DIOR-RSVG \cite{dior-rsvg}; the commonly used dataset is RSVG \cite{rsvg}; and the rare dataset is MAR20 \cite{wenqi2024mar20}. For the image scene task, the recommended datasets are EuroSAT \cite{eurosat}, RESISC45 \cite{resisc45}, WHU-RS19 \cite{whu-rs19}, SIRI-WHU\cite{siri-whu}, and AID \cite{aid-data}; the commonly used datasets are PatternNet \cite{patternnet}, RSI-CB256 \cite{rsi-cb}, METER-ML \cite{zhu2022meterml} and fMoW \cite{fmow}. \revii{Recent related benchmarks include MEET, which provides more than one million zoom-free samples for fine-grained geospatial scene classification~\cite{li2025meet}, and STAR, which supports scene graph generation in large-size satellite imagery~\cite{li2025star}.} For the image retrieval task, the recommended datasets are RSITMD \cite{rsitmd}, RSICD \cite{rsicd}, and UCM \cite{ucm}. At the same time, some VLGFMs have also released their own evaluation sets constructed for different specific capabilities.

\noindent$\bullet$
\textbf{Metrics.} For the image captioning task, the commonly used metrics include BiLingual Evaluation Understudy (BLEU), Recall-Oriented Understudy for Gisting Evaluation (ROUGE-L), Metric for Evaluation of Translation with Explicit ORdering (METEOR), and Consensus-based Image Description Evaluation (CIDEr). In VQA and image scene tasks, the primary metric used is Accuracy. For the visual grounding task, the commonly used metric is Accuracy@0.5, which considers a prediction accurate if the predicted bounding box overlaps with the ground-truth box by more than 0.5 Intersection over Union (IoU). The commonly used metric for the image retrieval task is the recall of top-K (R@K), which measures the ratio of queries that successfully retrieve the ground truth within the top K.

Additionally, experts can be employed to score the outcomes of responses. For instance, RSGPT \cite{hu2023rsgpt} utilizes experts to categorize the results of captioning tasks into four distinct levels. Since the advent of large language models, efforts have been made to leverage these models for scoring the outcomes of various vision-language multimodal tasks. For choice questions, the output of some models may not always meet the requirements due to imperfect instruction-following capabilities. \hl{Their responses may be off-topic, ambiguous, or formatted as meaningless numbers and special symbols, so an external judge is often needed to map free-form outputs to candidate options.} MMBench \cite{liu2023mmbench} employs ChatGPT \cite{gpt3} as a judge to identify the option most similar to the model's output and regard it as the model's prediction result.

\begin{table*}[!t]
\centering
\scriptsize
\caption{Comparison of mainstream remote sensing MLLMs on three image-captioning benchmarks. Values are reported in the comparison provided by RSCoVLM~\cite{li2025rscovlm}; a dash denotes an unreported result.}
\label{tab:caption-results}
\resizebox{0.72\linewidth}{!}{%
\begin{tabular}{l|l|ccccc}
\hline
Dataset & Model & BLEU-4 & METEOR & ROUGE-1 & ROUGE-L & CIDEr \\
\hline
\multirow{6}{*}{RSICD} & RSGPT~\cite{hu2023rsgpt} & 36.83 & 30.10 & -- & 53.34 & 102.94 \\
 & GeoChat~\cite{kuckreja2023geochat} & -- & 13.48 & 11.59 & -- & 12.39 \\
 & SkyEyeGPT~\cite{zhan2024skyeyegpt} & \textbf{59.99} & 35.35 & -- & 62.63 & 83.65 \\
 & RS-CapRet~\cite{silva2024large} & 43.40 & 37.00 & -- & \textbf{63.30} & \textbf{250.20} \\
 & EarthDial~\cite{soni2024earthdial} & -- & \textbf{56.18} & 33.77 & 27.61 & -- \\
 & GeoZero~\cite{wang2025geozero} & 35.35 & 52.31 & \textbf{60.16} & 54.47 & 114.82 \\
\hline
\multirow{8}{*}{UCM-Captions} & RSGPT~\cite{hu2023rsgpt} & 65.74 & 42.21 & -- & 78.34 & 333.23 \\
 & GeoChat~\cite{kuckreja2023geochat} & -- & 14.40 & -- & 13.22 & 14.27 \\
 & SkyEyeGPT~\cite{zhan2024skyeyegpt} & \textbf{78.41} & 46.24 & -- & 79.49 & 236.75 \\
 & RS-CapRet~\cite{silva2024large} & 67.00 & 47.20 & -- & 81.70 & 354.80 \\
 & RS-LLaVA~\cite{rsllava} & 72.84 & 47.98 & -- & 85.17 & 349.43 \\
 & EarthDial~\cite{soni2024earthdial} & -- & 51.42 & 40.00 & 34.15 & -- \\
 & RingMo-Agent~\cite{hu2025ringmoagent} & 77.63 & 51.79 & -- & \textbf{85.51} & 373.68 \\
 & GeoZero~\cite{wang2025geozero} & 75.72 & \textbf{83.75} & \textbf{84.60} & 83.75 & \textbf{393.57} \\
\hline
\multirow{6}{*}{Sydney-Captions} & RSGPT~\cite{hu2023rsgpt} & 62.23 & 41.37 & -- & 74.77 & 273.08 \\
 & GeoChat~\cite{kuckreja2023geochat} & -- & 10.63 & 12.00 & 11.26 & -- \\
 & SkyEyeGPT~\cite{zhan2024skyeyegpt} & \textbf{77.40} & 46.62 & -- & \textbf{77.74} & 181.06 \\
 & RS-CapRet~\cite{silva2024large} & 56.40 & 38.80 & -- & 70.70 & 239.20 \\
 & EarthDial~\cite{soni2024earthdial} & -- & 57.31 & 49.39 & 41.00 & -- \\
 & GeoZero~\cite{wang2025geozero} & 64.08 & \textbf{75.12} & \textbf{75.72} & 74.12 & \textbf{293.49} \\
\hline
\end{tabular}}
\end{table*}

\begin{table*}[!t]
\centering
\scriptsize
\caption{Comparison of mainstream remote sensing MLLMs on four VQA benchmarks. RSVQA-LR/HR~\cite{rsvqa}, XLRS-Bench~\cite{wang2025xlrsbench}, and LRS-VQA~\cite{luo2025lrsvqa} use different image scales and evaluation protocols; dashes indicate unreported results.}
\label{tab:vqa-results}
\begin{tabular}{l|cccc}
\hline
Model & RSVQA-LR & RSVQA-HR & XLRS-Bench & LRS-VQA \\
\hline
RSGPT~\cite{hu2023rsgpt} & 92.29 & 89.78 & -- & -- \\
GeoChat~\cite{kuckreja2023geochat} & 91.81 & 70.82 & 22.03 & 19.38 \\
SkyEyeGPT~\cite{zhan2024skyeyegpt} & 84.19 & 81.89 & -- & -- \\
EarthGPT~\cite{zhang2024earthgpt} & -- & 71.15 & -- & -- \\
VHM~\cite{pang2025vhm} & 89.12 & 74.35 & -- & 26.69 \\
RSUniVLM~\cite{liu2024rsunivlm} & 92.05 & \textbf{90.85} & -- & -- \\
RS-LLaVA~\cite{rsllava} & \textbf{93.04} & -- & -- & -- \\
LHRS-Bot-Nova~\cite{li2025lhrsbotnova} & -- & 88.45 & -- & -- \\
EarthDial~\cite{soni2024earthdial} & -- & 71.00 & -- & -- \\
RingMo-Agent~\cite{hu2025ringmoagent} & -- & 79.58 & -- & -- \\
ScoreRS~\cite{muhtar2025scorers} & -- & 81.55 & -- & -- \\
GeoZero~\cite{wang2025geozero} & -- & 88.92 & \textbf{48.10} & -- \\
ImageRAG~\cite{zhang2025imagerag} & -- & -- & -- & 27.92 \\
ZoomSearch~\cite{zhou2025zoomsearch} & -- & -- & -- & \textbf{32.92} \\
\hline
\end{tabular}
\end{table*}

\begin{table*}[!t]
\centering
\scriptsize
\caption{Comparison of mainstream remote sensing MLLMs on four region-level visual-grounding benchmarks: RSVG~\cite{rsvg}, DIOR-RSVG~\cite{dior-rsvg}, VRSBench~\cite{li2024vrsbench}, and AVVG~\cite{zhou2024geoground}. Values follow the comparison supplied by RSCoVLM~\cite{li2025rscovlm}.}
\label{tab:grounding-results}
\begin{tabular}{l|cccc}
\hline
Model & RSVG & DIOR-RSVG & VRSBench & AVVG \\
\hline
GeoChat~\cite{kuckreja2023geochat} & 2.04 & 24.05 & 11.52 & 0.28 \\
LHRS-Bot~\cite{muhtar2024lhrsbot} & 1.56 & 17.59 & 1.19 & 0.00 \\
SkySenseGPT~\cite{luo2024skysensegpt} & -- & 12.87 & -- & -- \\
LHRS-Bot-Nova~\cite{li2025lhrsbotnova} & -- & 31.51 & -- & -- \\
VHM~\cite{pang2025vhm} & -- & 56.17 & -- & -- \\
GeoGround~\cite{zhou2024geoground} & 26.65 & 77.73 & 66.04 & 21.58 \\
RSUniVLM~\cite{liu2024rsunivlm} & -- & 72.47 & 69.31 & -- \\
ScoreRS~\cite{muhtar2025scorers} & -- & 64.52 & -- & -- \\
GeoZero~\cite{wang2025geozero} & 50.04 & 79.43 & 73.32 & -- \\
RSCoVLM~\cite{li2025rscovlm} & \textbf{53.79} & \textbf{84.55} & \textbf{79.73} & \textbf{29.40} \\
\hline
\end{tabular}
\end{table*}

\begin{table*}[!t]
\centering
\scriptsize
\caption{Text-to-image retrieval results on RSICD and UCM-Captions, reproduced from Table IV of RS-CapRet~\cite{silva2024large}. R@K is recall at rank K, and mR is the reported mean of R@1, R@5, and R@10 in the text-to-image direction. Rows marked $\dagger$ were evaluated in the RS-CapRet setup; the RemoteCLIP values were collected in that paper from the original RemoteCLIP report~\cite{liu2024remoteclip}.}
\label{tab:retrieval-results}
\begin{tabular}{l|l|cccc}
\hline
Dataset & Model & R@1 & R@5 & R@10 & mR \\
\hline
\multirow{4}{*}{RSICD} & CLIP ViT-L (zero-shot)$^\dagger$~\cite{pmlr-v139-radford21a} & 5.03 & 19.03 & 30.25 & 18.10 \\
 & RemoteCLIP~\cite{liu2024remoteclip} & \textbf{14.73} & \textbf{39.93} & \textbf{56.58} & \textbf{37.08} \\
 & RS-CapRet$^\dagger$~\cite{silva2024large} & 9.83 & 30.17 & 47.43 & 29.14 \\
 & RS-CapRet (target-dataset tuned)$^\dagger$~\cite{silva2024large} & 10.25 & 31.62 & 48.53 & 30.13 \\
\hline
\multirow{4}{*}{UCM-Captions} & CLIP ViT-L (zero-shot)$^\dagger$~\cite{pmlr-v139-radford21a} & 10.76 & 46.00 & 73.33 & 43.37 \\
 & RemoteCLIP~\cite{liu2024remoteclip} & \textbf{17.71} & \textbf{62.19} & \textbf{93.90} & \textbf{57.93} \\
 & RS-CapRet$^\dagger$~\cite{silva2024large} & 15.52 & 57.24 & 88.76 & 53.84 \\
 & RS-CapRet (target-dataset tuned)$^\dagger$~\cite{silva2024large} & 16.10 & 56.29 & 90.76 & 54.38 \\
\hline
\end{tabular}
\end{table*}

\noindent \textbf{Quantitative synthesis.}
\revii{Tables \ref{tab:caption-results}--\ref{tab:retrieval-results} reveal four trends. First, image-captioning rankings remain metric- and dataset-dependent. SkyEyeGPT obtains the highest BLEU-4 on all three datasets, whereas GeoZero leads most METEOR, ROUGE-1, and CIDEr comparisons on UCM-Captions and Sydney-Captions; RS-CapRet remains strongest on RSICD ROUGE-L and CIDEr. Second, VQA performance is strongly tied to image scale and method specialization: RS-LLaVA leads RSVQA-LR, RSUniVLM leads RSVQA-HR, GeoZero leads XLRS-Bench, and the training-free ZoomSearch pipeline leads LRS-VQA. Third, the newer grounding-oriented systems show a large advantage over early conversational baselines, with RSCoVLM recording the best reported result on all four grounding benchmarks. Fourth, retrieval remains sensitive to domain alignment: RemoteCLIP outperforms zero-shot CLIP and the reported RS-CapRet variants on both RSICD and UCM-Captions, while dataset-specific tuning improves RS-CapRet only modestly and not uniformly across individual recall levels. Because the tables aggregate results reported under different model sizes, training corpora, and evaluation protocols, they should be read as a structured literature comparison rather than a controlled leaderboard.}

\subsection{Related Research Domains}

\noindent$\bullet$
\textbf{Vision geo-foundation models.} VGFMs are also called remote sensing pre-trained models. The primary difference between these models and VLGFM is that their input is purely visual, without any modeling of textual information. RVSA \cite{RVSA} trained a conventional vision transformer with 100 million parameters on RGB RSIs and developed a novel rotational resizing window attention mechanism to fine-tune the model for downstream tasks. Unlike MAE-based methods that infer the entire image from a few visible patches, RingMo \cite{ringmo} implemented a masked image modelling strategy in their pre-trained model, considering all image patches, whether masked or unmasked. The success of these preliminary studies demonstrates the significant potential of pre-trained models in remote sensing applications. Recently, foundational visual models for spectral \cite{spectralgpt}, multimodal \cite{guo2024skysense}, and synthetic aperture radar (SAR) \cite{l2024saratrx} images have been proposed, attracting significant attention in the VGFM field.

\noindent$\bullet$
\textbf{Remote sensing LLM-powered agent.}  The concept of an LLM-powered agent emphasizes leveraging the capabilities of LLMs to decompose tasks into multiple steps and then accomplish specific tasks by invoking appropriate tools. The most well-known agent framework currently is LangChain\footnote{https://www.langchain.com}, which allows developers to integrate large language models with external computation and data sources~\cite{wu2023visual,yang2023mm,shen2024hugginggpt,liu2023interngpt}. In the field of remote sensing interpretation, an LLM-powered agent \cite{guo2024remoteagent} would utilize pre-prepared toolkits, including object detection and classification algorithms, to complete visual-language multimodal tasks. Recently, RS-Agent \cite{xu2024rsagent} treats conversational VLGFMs, such as GeoChat and \revii{LHRS-Bot}, as tools, utilizing them to achieve RSI captioning functionality.

\section{Methods: A survey}

\begin{table*}[!t]
\centering 
\vskip 1mm
\caption{Train data of VLGFMs. Abbreviations IC, VQA, IS, Det., Seg., and VG in the table stand for image captioning, visual question answering, image scene classification, object detection, segmentation, and visual grounding, respectively (If more than 1 models use the dataset to train a task, it is highlighted in \textcolor{blue}{blue}. If more than 2 models use it, it is highlighted in \textcolor{red}{red}).}
\vskip 1mm
\resizebox{\linewidth}{!}{
\begin{tabular}{c|cccccc}
\hline
Model & IC & VQA & IS & Det./Seg. & VG  & Others \\ 
\hline
\multirow{3}{*}{RemoteCLIP} & \multirow{3}{*}{\textcolor{red}{RSICD}, \textcolor{red}{RSITMD},} & \multirow{4}{*}{-} & \multirow{4}{*}{-} & \textcolor{red}{DOTA}, \textcolor{red}{DIOR}, \textcolor{blue}{HRRSD}, \textcolor{blue}{RSOD}, & \multirow{4}{*}{-} &  \multirow{4}{*}{-} \\ 
\multirow{3}{*}{\cite{liu2024remoteclip}} & \multirow{3}{*}{\textcolor{red}{UCM-Captions}} & &  &LEVIR, HRSC, \textcolor{blue}{VisDrone}, &  & \\
 &  & &  & AU-AIR, S-Drone, CAPRK,&  & \\
 &  & &  & Vaihingen, Potsdam, iSAID, \textcolor{blue}{LoveDA}&  & \\
\hline
GeoRSCLIP & \multirow{2}{*}{-} & \multirow{2}{*}{-} & \textcolor{red}{fMoW}, BEN, & \multirow{2}{*}{-} & \multirow{2}{*}{-} &  \multirow{2}{*}{RS5M} \\ 
\cite{zhang2024rs5m} &  & & \textcolor{blue}{MillionAID} & &  & \\
\hline
RSGPT & \multirow{2}{*}{-} & \multirow{2}{*}{-} & \multirow{2}{*}{-} & \multirow{2}{*}{-} & \multirow{2}{*}{-} &  \multirow{2}{*}{RSICap} \\ 
\cite{hu2023rsgpt} &  & &  & &  & \\
\hline
GeoChat & \multirow{2}{*}{-} & \textcolor{red}{FloodNet}, & \textcolor{blue}{RESISC45} & \textcolor{red}{DOTA}, \textcolor{red}{DIOR}, \textcolor{red}{FAR1M}, & \multirow{2}{*}{-} &  \multirow{2}{*}{-} \\ 
\cite{kuckreja2023geochat} &  & \textcolor{red}{RSVQA-LR}&  & SAMRS&  & \\
\hline
GRAFT & \multirow{2}{*}{-} & \multirow{2}{*}{-} & \multirow{2}{*}{-} & \multirow{2}{*}{-} & \multirow{2}{*}{-} & NAIP, \\ 
\cite{mall2023remote} &  & &  & &  & Sentinel-2\\
\hline
SkyCLIP & \multirow{2}{*}{-} & \multirow{2}{*}{-} & \multirow{2}{*}{-} & \multirow{2}{*}{-} & \multirow{2}{*}{-} & \multirow{2}{*}{SkyScript} \\ 
\cite{wang2024skyscript} &  & &  & &  & \\
\hline
\multirow{4}{*}{SkyEyeGPT} & \textcolor{red}{RSICD}, \textcolor{red}{RSITMD}, & \multirow{2}{*}{ERA-VQA,} & \multirow{5}{*}{-} & \multirow{5}{*}{-} & \multirow{3}{*}{\textcolor{blue}{RSVG},} &  \multirow{2}{*}{DOTA-Conversa} \\ 
\multirow{4}{*}{\cite{zhan2024skyeyegpt}} & \textcolor{red}{UCM-Captions}, & \multirow{2}{*}{\textcolor{red}{RSVQA-LR},} &  & & \multirow{3}{*}{\textcolor{red}{DIOR-RSVG},} & \multirow{2}{*}{DIOR-Conversa} \\
 & \textcolor{red}{Sydney-Captions}, &\multirow{2}{*}{\textcolor{red}{RSVQA-HR},} &  & & \multirow{3}{*}{RSPG} & \multirow{2}{*}{UCM-Conversa}\\
 & \textcolor{red}{NWPU-Captions}, & \multirow{2}{*}{\textcolor{blue}{RSIVQA}} &  & &  & \multirow{2}{*}{Sydney-Conversa}\\
 & CapERA & &  & &  & \\
\hline
\multirow{4}{*}{EarthGPT} & \textcolor{red}{RSICD}, \textcolor{red}{RSITMD}, & \multirow{2}{*}{\textcolor{red}{FloodNet},} & \multirow{2}{*}{DSCR, EuroSAT,} & \textcolor{red}{DOTA}, \textcolor{red}{DIOR}, \textcolor{red}{FAR1M}, NWPUVHR10, & \multirow{5}{*}{\textcolor{red}{DIOR-RSVG}} & \multirow{5}{*}{MMRS-1M} \\ 
\multirow{4}{*}{\cite{zhang2024earthgpt}} & \textcolor{red}{UCM-Captions}, & \multirow{2}{*}{\textcolor{red}{RSVQA-LR},} & \multirow{2}{*}{\textcolor{red}{UCM}, WHU-RS19,} & \textcolor{blue}{HRRSD}, \textcolor{blue}{RSOD}, UCAS-AOD, \textcolor{blue}{VisDrone}, &  & \\
 & \textcolor{red}{Sydney-Captions}, &\multirow{2}{*}{\textcolor{red}{RSVQA-HR},} & \multirow{2}{*}{RSSCN7, \textcolor{blue}{RESISC45},}  & AIR-SARShip-2.0, SSDD, HRISD, HIT-UAV, &  & \\
 & \textcolor{red}{NWPU-Captions} & \multirow{2}{*}{CRSVQA} & \multirow{2}{*}{FGSCR-42} & Sea-shipping, Infrared-security, Aerial-mancar,&  & \\
 &  &  &  & HIT-UAV, Double-light-vehicle, Oceanic ship &  & \\
\hline
\multirow{2}{*}{LHRS-Bot} & \textcolor{red}{RSICD},\textcolor{red}{RSITMD}, & \multirow{2}{*}{\textcolor{red}{RSVQA-LR}} & \textcolor{red}{RSITMD}, \textcolor{red}{NWPU}, & \multirow{3}{*}{-} & \multirow{2}{*}{\textcolor{blue}{RSVG}} & \multirow{2}{*}{LHRS-Align,} \\ 
\multirow{2}{*}{\cite{muhtar2024lhrsbot}}&\textcolor{red}{UCM-Caption}&\multirow{2}{*}{\textcolor{red}{RSVQA-HR}}&\textcolor{blue}{METER-ML},\textcolor{red}{UCM},&&\multirow{2}{*}{\textcolor{red}{DIOR-RSVG}}& \multirow{2}{*}{LHRS-Instruct}\\
 & \textcolor{red}{NWPU-Captions} &  & \textcolor{red}{fMoW} & &  & \\
\hline
\multirow{3}{*}{RS-CapRet} & \textcolor{red}{RSICD}, & \multirow{4}{*}{-} & \multirow{4}{*}{-} & \multirow{4}{*}{-} & \multirow{4}{*}{-} &  \multirow{4}{*}{-} \\ 
\multirow{3}{*}{\cite{silva2024large}} & \textcolor{red}{UCM-Captions}, & &  & &  & \\
& \textcolor{red}{Sydney-Captions}, & &  & &  & \\
& \textcolor{red}{NWPU-Captions} & &  & &  & \\
\hline
\multirow{2}{*}{H$^2$RSVLM} & \multirow{3}{*}{-} & \multirow{3}{*}{\textcolor{red}{RSVQA-LR}} & \textcolor{red}{fMoW}, \textcolor{red}{RSITMD}, &  \textcolor{red}{DOTA}, \textcolor{red}{FAIR1M}, \textcolor{blue}{LoveDA},  & \multirow{3}{*}{\textcolor{red}{DIOR-RSVG}} & CVUSA, CVACT,\\
\multirow{2}{*}{\cite{pang2024h2rsvlm}} &  & & \textcolor{blue}{MillionAID}, \textcolor{red}{NWPU}, & MSAR, GID, FBP, &  & BANDON, \\
 &  & & \textcolor{blue}{METER-ML}, \textcolor{red}{UCM} & DeepGlobe, CrowdAI &  & MtS-WH \\
\hline
RS-LLaVA & \textcolor{red}{UCM-Captions}, & \textcolor{red}{RSVQA-LR}, & \multirow{2}{*}{\textcolor{red}{UCM}} & \multirow{2}{*}{-} & \multirow{2}{*}{-} & \multirow{2}{*}{-}\\
\cite{rsllava} & UAV & \textcolor{blue}{RSIVQA} & \multirow{2}{*}{-} &  &  &  \\
\hline
\multirow{2}{*}{SkySenseGPT} & & \textcolor{red}{FloodNet}&\textcolor{red}{RSITMD},&\textcolor{red}{DOTA}&& \multirow{3}{*}{RSG}\\
\multirow{2}{*}{\cite{luo2024skysensegpt}} &  & \textcolor{red}{RSVQA-LR} & \textcolor{red}{UCM}, & \textcolor{red}{DIOR}& & \\
&&Earth-VQA&\textcolor{red}{NWPU}&\textcolor{red}{FAIR1M}&& \\
\hline
RSCoVLM & VHM-Instruct, & VHM-Instruct, & \textcolor{blue}{VHM-Instruct}, & \multirow{2}{*}{\textcolor{red}{DOTA}} & refGeo, &  \multirow{2}{*}{LLaVA-OneVision} \\ 
\cite{li2025rscovlm} & GeoChat & GeoChat & GeoChat & & TEOChatlas & \\
\hline
\multirow{4}{*}{GeoZero} &  & \multirow{2}{*}{VHM-Instruct,} & \textcolor{red}{fMoW}, \textcolor{blue}{EuroSAT} & \multirow{5}{*}{-} & \multirow{2}{*}{\textcolor{red}{RSVG},} &  \multirow{5}{*}{VHM-Instruct} \\ 
\multirow{4}{*}{\cite{wang2025geozero}} &  & \multirow{2}{*}{\textcolor{red}{RSVQA-LR},} & \textcolor{blue}{WHU-RS19}, & & \multirow{2}{*}{\textcolor{red}{DIOR-RSVG},} &  \\
 & SkyEye968k, &\multirow{2}{*}{\textcolor{red}{RSVQA-HR},} & \textcolor{red}{RESISC45}, & & \multirow{2}{*}{VRSBench} & \\
 & VRSBench & \multirow{2}{*}{VRSBench} & NASC-TG2, AID & & \multirow{2}{*}{VHM-Instruct} & \\
 & & & \textcolor{blue}{VHM-Instruct} & &  & \\
\hline
\end{tabular}}
\label{Table:train-data}
\end{table*}

\textbf{Overview.} In this section, we systematically explore how data is gathered and utilized to train these models (Section \ref{sec:datacollection}), examine the architectural choices and modifications made to enhance their performance (Section \ref{sec:architecture}), and precisely define their capabilities by the variety of tasks they can handle (Section \ref{sec:capabilities}). \hl{To avoid an arbitrary selection of examples, the representative works discussed in detail are selected according to four criteria: direct relevance to VLGFMs, release of models or datasets, coverage of important capabilities or benchmarks, and clear influence on subsequent remote sensing vision-language studies. Other relevant works are retained in the tables and in the continuously maintained repository.}

\begin{table*}
\centering 
\vskip 1mm
\caption{Data Pipeline of VLGFMs.}
\vskip 1mm
\resizebox{\linewidth}{!}{
\begin{tabular}{c|l}
\hline
Model & Data pipeline \\ 
\hline
\multirow{3}{*}{RemoteCLIP} & \textbf{Retrieval Data}: use it directly. \textbf{Detection Data}: generate 5 distinct captions that describe the objects in the image \\
\multirow{3}{*}{\cite{liu2024remoteclip}} &  (the first 2 captions are generated according to the target location, and the remaining 3 captions are generated by\\
& considering the number of different object categories). \textbf{Segmentation Data}: first convert it to boxes, then use the \\
& same pipeline as the detection data.\\
\hline
\multirow{3}{*}{RS5M} & \textbf{PUB11}: \revii{use regular expressions to identify image--text pairs containing remote-sensing keywords and fastdup to detect invalid} \\
\multirow{3}{*}{\cite{zhang2024rs5m}} &  \revii{images and duplicates. Then clean the dataset using CLIP scores and an RS image classifier.} \textbf{RS3}:  \\
& generate 20 candidate captions per image using tuned BLIP2 and rank the top 10 results using CLIP ViT-H/14.  \\
& Then, re-rank these top 10 results using CLIP Resnet50x64 to obtain the top 5 captions. \\
\hline
\multirow{4}{*}{RSGPT} & \textbf{RSICap}: 5 RS experts annotate 2,500 images from the detection dataset, focusing on (1) image attributes like   \\
\multirow{4}{*}{\cite{hu2023rsgpt}}& type,  and resolution; (2) object details such as quantity, color, shape, size, and position; (3) annotate the overall \\
&  scene before specific objects. \textbf{RSIEval}: \revii{for captioning, follow RSICap with one caption per image; for VQA,} \\
&  \revii{produce nine question--answer pairs per image, categorized as object-related, image-related, scene-}\\
& related, and reasoning-related questions. \\
\hline
\multirow{5}{*}{GeoChat} & \textbf{Attribute extraction}: Objects' categories are sourced from the SAMRS dataset. Objects' colors are obtained by \\
\multirow{5}{*}{\cite{kuckreja2023geochat}} & K-Means. Objects' relative sizes are small, normal, or large. Objects' relative positions are assessed using a 3x3  \\
&  grid overlay on images, and the relationship of objects' bounding boxes determines their relations. \textbf{Expression}\\

& \textbf{Generation}: employ predefined textual templates based on objects' attributes to emulate natural language  \\
& expressions. \textbf{Instruction Generation}: generated referring expressions are input into an LLM, which outputs 3\\
&  kinds of question-answer pairs, i.e. grounding image description, referring expression, and region captioning. \\
\hline
\multirow{4}{*}{GRAFT} & \textbf{Ground Images}: We sourced outdoor ground images with precise geo-tags from diverse regions on Flickr, \\
\multirow{4}{*}{\cite{mall2023remote}} & ensuring a uniform location distribution avoiding duplicates, and filtering out indoor photos with a classifier.\\
& \textbf{Satellite Images}: Satellite images are sampled with centres at the geotags of ground images, with each assigned \\
& all non-duplicate ground images within its bounds. Overlaps between satellite images are minimized (at least 112  \\
& pixels apart). For any location with more than 25 ground images, a random subset of 25 is selected. \\
\hline
\multirow{4}{*}{SkyScript} & Images are from Google Earth Engine (GEE), and the geographic features are gathered from OpenStreetMap  \\
\multirow{4}{*}{\cite{wang2024skyscript}} & (OSM). Connect images with appropriate OSM semantics using a two-stage tag classification approach with \\
& CLIP tag embeddings. Converting tags into a caption by first connecting the key and the value with a connecting \\
& word (e.g., "of", "is") and then connecting multiple tags with commas or "and".  Filtering out uncorrelated image\\

& -text pairs by CLIP model to derive embeddings and compute cosine similarity.\\

\hline
\multirow{6}{*}{SkyEyeGPT} & \textbf{Single-task Image-text Instruction}: integrate 5 image captioning, 1 video captioning, and 4  VQA datasets. \\
\multirow{6}{*}{\cite{zhan2024skyeyegpt}} & Following the miniGPT-v2, create an phrase grounding dataset. \textbf{Multi-task Conversation Instruction}: mix the\\
&  corresponding captioning and VQA datasets to get UCM-Conversa and Sydney-Conversa instruction. Using the \\
& DIOR-RSVG and DIOR to construct DIOR-Conversa instruction which contains visual grounding, phrase \\
& grounding, and referring expression generation tasks. Leverage RSIVQA and the DOTA to build a conversation \\
& instruction, DOTA-Conversa, that includes VQA and phrase grounding tasks. Data is manually verified and \\
& selected by humans to ensure the high quality of the instruction.\\
\hline
EarthGPT & Convert 10 classification datasets, 5 image captioning datasets,  4 VQA datasets, 17 object detection datasets, and \\
\cite{zhang2024earthgpt} & 1 visual grounding dataset into instructions using prompt templates.\\
\hline
\multirow{8}{*}{LHRS-Bot} & \textbf{LHRS-Align}: images are from GEE, and the geographic features are gathered from OSM. Filtering out irrelevant \\
\multirow{8}{*}{\cite{muhtar2024lhrsbot}} & keys and balance semantic information. Using Vicuna-v1.5 to generate image captions by summarising the key\\
& -value tags of the corresponding images. \textbf{LHRS-Instruct}: For public RS caption dataset, a rigorous data cleaning  \\
& process is implemented, including deduplication, caption length computation, and relevance assessment via CLIP \\
& score. Several in-context examples with the five captions per image prompt Vicuna-v1.5 to generate object-focused  \\
& conversations.  The most relevant conversations are filtered using predefined criteria and expert evaluations.  For \\
& LHRS-Align, selecting the attribute-rich RS images from the LHRS-Align dataset, bounding boxes for objects \\
& within these images are calculated using OSM database spatial coordinates. This enriched data prompts GPT-v4  \\
& to create complex instruction data featuring visual reasoning, detailed descriptions, and object location and count \\
& conversations. The final step involves a manual cleaning of the generated results. \\
\hline
\multirow{9}{*}{H$^2$RSVLM} & \textbf{HqDC-1.4M}: designed appropriate prompts to guide the Gemini-Vision
model to provide detailed descriptions of  \\ 
\multirow{9}{*}{\cite{pang2024h2rsvlm}} & remote sensing images, including the type and resolution of the images, as well as information on the attributes, \\
& quantities, colors, shapes, and spatial positions of objects in
the images. \textbf{HqDC-Instruct}: combined the \\
& aforementioned descriptions with the original object detection annotations of the dataset and input them into the \\
& language-only Gemini. Used prompts and few-shot examples to guide Gemini in generating multi-turn conversation \\
& and reasoning data. \textbf{RSSA}: Designed question-answer pairs based on two RS object detection datasets, some of  \\
& which questions cannot be answered using the information in the images alone. \textbf{RS-Specialized-Instruct}: A remote  \\
& sensing-specific instruction-following dataset created using some existing remote sensing datasets. \textbf{RS-ClsQaGrd-}\\
& \textbf{Instruct}: a multi-task instruct dataset constructed by
public remote sensing datasets, covering tasks such as scene \\
& classification, VQA, and
visual grounding.\\
\hline
RS-LLaVA & \multirow{2}{*}{Convert 2 image captioning datasets and 2 VQA datasets into instructions using prompt templates.}  \\ 
\cite{rsllava} &  \\
\hline
SkySenseGPT & \textbf{FIT-RS}: enhance a scene graph generation remote sensing dataset, which contains complex relationships between\\
\cite{luo2024skysensegpt} & remote sensing objects. \\
\hline
GeoZero \cite{wang2025geozero} & \textbf{GeoZero-Raw}: Utilize the existing 13 RS vision-language datasets.\\
\hline
\end{tabular}}
\label{Table3}
\end{table*}

\noindent \textbf{Motivation of survey organization.} \hl{We organize the method section around data, architecture, and capability because these three dimensions explain most differences among current VLGFMs. Data construction determines the semantic fidelity, geographic coverage, and task supervision available to the model; architecture determines how visual and textual information are aligned or generated; and capability analysis connects these technical choices to observable performance.} We believe that the current emphasis in VLGFM research should be more on data rather than fundamentally altering the architecture of general large models and training them from scratch. There are several reasons. (1) The available image-text data in remote sensing is significantly less than in computer vision. This gap makes it infeasible to support the training of a remote sensing image-text alignment model from scratch. Therefore, constructing high-quality, large-scale remote sensing image-text datasets remains a top priority for VLGFM research in the foreseeable future. (2) Training VLGFMs from scratch requires substantial time and resources. Without significant resource investment, any short-term lead is likely to be quickly overtaken by the rapidly evolving general-purpose multimodal models. (3) VLGFMs represent the application of large general models in the remote sensing domain, falling under the category of specialized artificial intelligence. The key is to leverage the extensive prior knowledge in the remote sensing field to address specific, real-world problems. Thus, fine-tuning networks using high-quality remote sensing image-text datasets to enable the model to generalize and apply knowledge effectively is a practical approach.

\subsection{Data Pipeline}
\label{sec:datacollection}

\revii{VLGFMs are commonly adapted with remote sensing image--text pairs. Existing pipelines combine four recurring strategies: expert annotation, template conversion of task labels, model-generated language, and alignment with external geographic or volunteered data. The following discussion first describes the two broad acquisition routes and then compares these four supervision strategies directly.}

\noindent$\bullet$
\textbf{Data collection from scratch.} As shown in Table \ref{Table:train-data}, three VLGFMs have built their training datasets from scratch rather than utilizing existing ones. RSGPT \cite{hu2023rsgpt} prioritizes the quality of training samples over their quantity. Hence, it employs five experts to annotate images, producing high-quality captions manually. As a result, the model achieves satisfactory performance with fine-tuning on only 2,500 remote sensing image-text pairs. GRAFT \cite{mall2023remote} \hl{uses} the geo-tag information of images to link ground-level image captions collected from social media with RSIs. This approach enables the acquisition of many remote sensing image-text pairs without the need for manual annotation of RSI captions. SkyScript \cite{wang2024skyscript} utilizes a well-known open-source collaborative project, OpenStreetMap\footnote{https://www.openstreetmap.org} (OSM), to construct image captions. OSM allows users to contribute and edit geospatial data, such as road networks, trails, cafes, railway stations, etc. It provides a comprehensive and freely accessible world map and detailed user annotations. SkyScript filters these annotations and combines various tags to generate high-quality image captions. \revii{In LHRS-Bot} \cite{muhtar2024lhrsbot}, the LHRS-Align also acquires tags of geographic features from OSM and generates image captions based on these tags using \revii{the language-only model Vicuna-v1.5}. More details can be found in Table \ref{Table3}.

\begin{figure*}[!h]
\centering
\includegraphics[width=16cm]{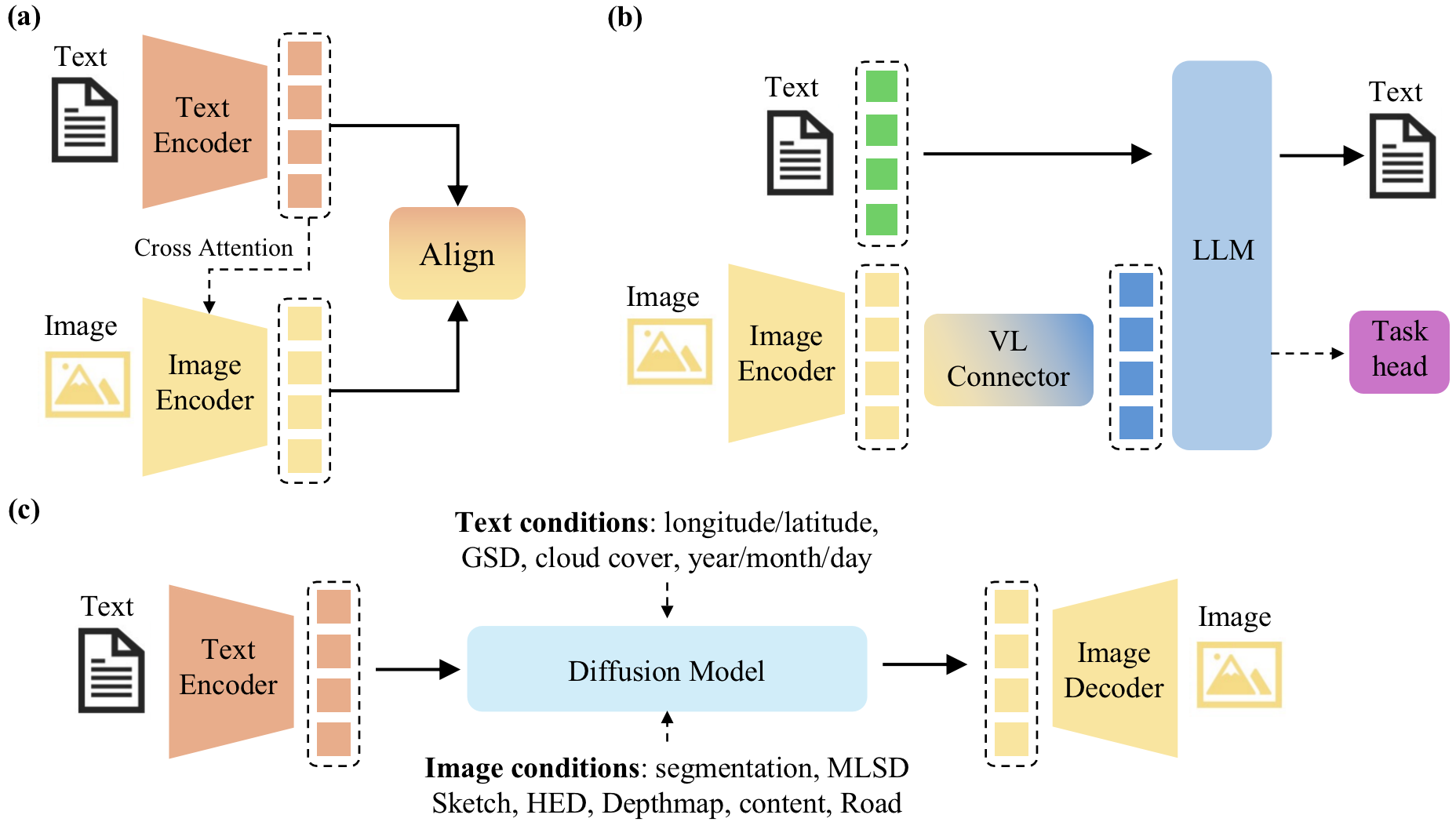}
\caption{Architecture of representative VLGFMs: (a) Contrastive, (b) Conversational, (c) Generative.}
\label{fig:arch}
\end{figure*}

\noindent$\bullet$
\textbf{Data enhancement using existing datasets.} Most works choose to modify existing remote sensing datasets, which is significantly less labour-intensive compared to building datasets from scratch. Consequently, researchers can focus more on improving the model's structures. RemoteCLIP \cite{liu2024remoteclip} was the first to propose using text templates to convert the most common object detection annotations in the field of remote sensing interpretation into image captions, generating a large number of remote sensing image-text pairs suitable for training large models. Subsequently, EarthGPT \cite{zhang2024earthgpt} also adopted the same approach. RS5M \cite{zhang2024rs5m}, which currently contains the largest image-text pairs, includes 5 million RSIs. It employs the general multimodal model BLIP2 \cite{blip2} to generate image captions and then uses CLIP \cite{pmlr-v139-radford21a} to select the top 5 highest-scoring captions. GeoChat \cite{kuckreja2023geochat} constructs a remote sensing image-text pair dataset based on the segmentation dataset generated using SAM \cite{SAM} technology. It first uses the target's color, size, and location to build basic expressions and then inputs these expressions into a language-only LLM to generate question-answer pairs. SkyEyeGPT \cite{zhan2024skyeyegpt} constructs a multi-task conversation instruction dataset by combining an object detection dataset with a VQA dataset built using the object detection data. This approach supports training for both VQA and visual grounding tasks. For more details, please refer to Table \ref{Table3}.

\noindent \textbf{Trade-off analysis of data strategies.}
\revii{Table~\ref{tab:data-strategy} separates the four strategies that were previously compressed into a narrative comparison. No strategy dominates every criterion: expert annotation maximizes semantic fidelity, while automated conversion and external-data alignment increase scale and geographic breadth at the cost of distinct noise and bias modes.}

\begin{table*}[!t]
\centering
\scriptsize
\caption{Comparison of four recurring VLGFM data-construction strategies.}
\label{tab:data-strategy}
\setlength{\tabcolsep}{4pt}
\renewcommand{\arraystretch}{1.18}
\begin{tabularx}{\linewidth}{>{\RaggedRight\arraybackslash}p{3.0cm}>{\RaggedRight\arraybackslash}p{2.6cm}>{\RaggedRight\arraybackslash}p{3.2cm}>{\RaggedRight\arraybackslash}X>{\RaggedRight\arraybackslash}p{3.5cm}}
\toprule
\textbf{Strategy and examples} & \textbf{Scale / cost / fidelity} & \textbf{Primary benefit} & \textbf{Main geo-specific risks} & \textbf{Recommended controls} \\
\midrule
Expert annotation (RSICap/RSGPT) & Small / high / high & Precise, auditable captions and reasoning supervision & Limited regional, seasonal, sensor, and annotator diversity; weak long-tail coverage & Stratified multi-region sampling; dual annotation; adjudication; uncertainty labels \\
Template conversion (RemoteCLIP, EarthGPT; GeoChat hybrid) & Large / low--medium / label-exact & Scalable conversion of class, box, mask, and relation labels & Template artifacts; label leakage; simplified relations; inherited class and geographic bias & Diverse templates; source-disjoint tests; coordinate and relation validation \\
Model-generated language (RS5M, SkyEyeGPT; GeoChat hybrid) & Very large / medium / variable & Broad instruction and dialogue coverage & Hallucinated objects or relations; teacher-model bias; verbosity; inconsistent coordinates & Visual-grounding checks; score-based filtering; human audits; consistency tests \\
External geo-data alignment (GRAFT, SkyScript, LHRS-Align) & Large / low--medium / weak & Broad geographic coverage and explicit geographic semantics & Temporal or footprint mismatch; incomplete VGI tags; participation bias; ground-to-overhead mismatch & Date and footprint matching; geographic balancing; tag confidence; cross-source validation \\
\bottomrule
\end{tabularx}
\end{table*}

\begin{table*}[!t]
\centering
\scriptsize
\caption{Subset of the LHRS-Bot alignment-data ablation reported in Table 4 of~\cite{muhtar2024lhrsbot}. Only the first-stage alignment dataset is changed, while the remaining training and evaluation settings are held fixed. Dataset size is the number of image--text pairs, and all task values are accuracy (\%).}
\label{tab:lhrs-ablation}
\begin{tabular}{l|c|cc|cc}
\hline
Alignment data & Pairs & RSVQA-LR & RSVQA-HR & RSVG & DIOR-RSVG \\
\hline
RS5M & 5.0M & 78.36 & 91.65 & 70.58 & 87.12 \\
SkyScript & 1.5M & 88.79 & 91.24 & 71.25 & 85.05 \\
LHRS-Align & 1.15M & \textbf{89.19} & \textbf{92.55} & \textbf{73.45} & \textbf{88.10} \\
\hline
\end{tabular}
\end{table*}

\noindent \textbf{Observed effect on downstream accuracy.}
\revii{The controlled LHRS-Bot comparison in Table~\ref{tab:lhrs-ablation} shows that the 1.15M-pair LHRS-Align corpus outperforms both the 5M-pair RS5M corpus and the 1.5M-pair SkyScript corpus on the two reported VQA and two grounding datasets. For example, compared with RS5M, LHRS-Align improves RSVQA-LR from 78.36 to 89.19 and RSVG from 70.58 to 73.45. Because the experiment changes the first-stage alignment corpus as a whole, it supports the value of domain-specific data but does not isolate the effects of caption generation, imagery, and data cleaning.}
\subsection{Architecture}
\label{sec:architecture}

VLGFMs can be briefly classified into three categories based on their input-output mechanisms. Contrastive VLGFMs \cite{liu2024remoteclip, zhang2024rs5m, mall2023remote, wang2024skyscript} take both text and images as inputs and produce a similarity measure between them. This similarity is crucial for image-text retrieval and zero-shot scene classification applications. Conversational VLGFMs \cite{hu2023rsgpt,kuckreja2023geochat,zhan2024skyeyegpt,zhang2024earthgpt,muhtar2024lhrsbot,silva2024large,pang2024h2rsvlm} also process text and images as inputs but output textual responses, leveraging the powerful language capabilities of LLMs to support tasks such as captioning and visual question answering. Additionally, dense prediction tasks can be supported by appending task-specific heads to the LLM. On the other hand, generative VLGFMs \cite{khanna2023diffusionsat,tang2024crsdiff} are conditioned on either text or images, producing images as outputs. These models utilize conditional diffusion processes to generate controlled synthetic RSIs. This section will provide a detailed exposition of the model architectures for these three types of VLGFMs.

\subsubsection{Contrastive VLGFMs}

A typical contrastive VLGFM comprises two main components: an image encoder and a text encoder. A schematic of this architecture is depicted in Figure \ref{fig:arch}. Image encoder uses a convolutional neural network or a vision transformer to encode images into a feature vector. Parallel to the vision encoder, the text encoder (typically a BERT model) processes textual descriptions into another feature vector. The core idea behind contrastive VLGFM is to align the image and text embeddings in a shared multidimensional space. During training, contrastive VLGFM uses a contrastive learning objective, specifically a noise contrastive estimation technique variant. The model is trained to maximize the similarity between positive image-text pairs and minimize the similarity between negative pairs. The large-scale training allows contrastive VLGFM to understand various visual concepts described in natural language robustly. CLIP \cite{pmlr-v139-radford21a} is one of the pioneering models for vision-language pre-training. It employs the InfoNCE loss to align text and images in the feature space. Benefiting from a dataset comprising 400 million image-text pairs, CLIP has achieved significant success in various multimodal downstream tasks. BLIP \cite{blip} is the first multimodal model that simultaneously addresses both understanding and generation of image-text content, which is jointly pre-trained on three vision-language tasks: image-text contrastive learning, image-text matching, and image-conditioned language modelling. 

In the domain of earth observation, some outstanding vision-language pre-training models have also emerged. RemoteCLIP \cite{liu2024remoteclip} is the first true contrastive VLGFM. It utilizes existing remote sensing retrieval, detection, and segmentation datasets to construct image-text sample pairs, supporting model training. GeoRSCLIP is a vision-language pre-training model released alongside the RS5M dataset \cite{zhang2024rs5m}. The authors did not provide extensive details about it in the paper. GRAFT \cite{mall2023remote} is a significant attempt to address the issue of insufficient samples in remote sensing image-text data. It utilizes a vast collection of ground-level images from Flickr\footnote{https://www.flickr.com} as a bridge, pairing RSIs that correspond to the geographic coordinates of these ground images with the titles of the ground images. This approach allows for the construction of a large-scale dataset of remote sensing image-text pairs without the need for additional manual annotation. SkyScript \cite{wang2024skyscript} utilizes the ground sample distance (GSD) of OSM data to annotate images appropriately. For instance, house numbers cannot be discerned from RSIs regardless of their resolution, and utility poles are not visible in RSIs with a GSD of 10. However, these contrastive VLGFMs are primarily data-centric efforts, with no significant modifications made to the model structures.

\subsubsection{Conversational VLGFMs}

A typical conversational VLGFM comprises three main components: a pre-trained visual encoder, a pre-trained LLM, and a modality interface that connects them. Drawing an analogy with human sensory and cognitive processes, the visual encoder functions similarly to the human eye, receiving and preprocessing optical signals. The LLM, akin to the human brain, can understand these processed signals and perform reasoning tasks. The interface between them serves to align the visual and linguistic modalities. A schematic of this architecture is depicted in Figure \ref{fig:arch}. In this section, we will describe each of these components in detail.

\noindent$\bullet$
\textbf{Visual encoder.} The visual encoder compresses images into more compact representations. Training this network from scratch is time-consuming; therefore, a common approach is to utilize the image encoder from an already text-aligned contrastive VLGFM. Aligning the visual encoder with textual semantics is advantageous for subsequent integration with LLMs.

Table \ref{Table0} presents the visual encoders commonly used in current VLGFM implementations. Besides the standard CLIP image encoder \cite{pmlr-v139-radford21a}, some studies have explored other variants. For instance, RSGPT and SkyEyeGPT employ the EVA-CLIP encoder \cite{sun2023evaclip}, which is trained with improved training techniques. Additionally, EarthGPT utilizes two encoders simultaneously and introduces the ConvNext encoder \cite{convnext}, a convolution-based architecture designed to exploit higher resolutions and multi-level features.

\noindent$\bullet$
\textbf{Pre-trained LLM.} Training an LLM from scratch entails prohibitive computational costs for most individuals, making it more effective and practical to begin with a pre-trained one. Through pre-training on vast amounts of data collected from various places, the LLM acquires rich world knowledge and demonstrates strong capabilities in generalization and reasoning. Table \ref{Table0} summarizes commonly used and publicly available LLMs, all of which fall under the category of causal decoders \cite{gpt3}. The LLAMA series \cite{llava} and the Vicuna family \cite{vicuna} represent some of the most notable open-source LLMs that have garnered considerable attention within the academic community. It is important to note that these LLMs were primarily pre-trained on English corpora, which limits their multilingual capabilities, such as for Chinese. Additionally, increasing the parameter size of LLMs also brings additional benefits, akin to increasing input resolution. Specifically, RSGPT \cite{hu2023rsgpt} found that simply expanding the Vicuna from 7B to 13B parameters can significantly enhance performance across various benchmarks.

\noindent$\bullet$
\textbf{Vision-language connector.} Since LLMs are limited to text perception, bridging the gap between natural language and visual modalities is necessary. However, training large multimodal models in an end-to-end manner would be costly. A more practical approach is introducing learnable vision-language connectors between pre-trained visual encoders and LLMs. The module efficiently projects image information into a space that LLMs can understand. Based on the fusion method between visual and language information, these connectors can generally be divided into token-level and feature-level connectors.

For token-level connectors, the output of the encoder is transformed into tokens and concatenated with text tokens before being fed into the LLM. A common and feasible solution is to use a set of learnable query tokens to extract information in a query-based manner, initially implemented with BLIP-2 \cite{blip2} and subsequently adopted by various works on VLGFMs \cite{hu2023rsgpt}. This query-based approach compresses visual tokens into a smaller number of representation vectors. In contrast, some methods employ multi-layer perception (MLP) to bridge the modal gap. For instance, the LLaVA series only use linear layers \cite{zhan2024skyeyegpt,zhang2024earthgpt,silva2024large} or two-layer MLPs \cite{kuckreja2023geochat,pang2024h2rsvlm} to project visual tokens and align feature dimensions with word embeddings.

For feature-level connectors, it is required to incorporate additional modules to facilitate deep interaction and integration between textual features and visual features. For instance, Flamingo \cite{flamingo} introduces additional cross-attention layers between the frozen transformer layers of the LLM, thereby utilizing external visual cues to enhance language features. Similarly, LHRS-Bot \cite{muhtar2024lhrsbot} attempts to introduce a set of learnable queries for each level of image features. These queries aim to encapsulate the semantic information of each level of visual features through a series of stacked cross-attention and MLP layers.

\noindent$\bullet$
\textbf{Pre-training.} Pre-training primarily aims to align different modalities and learn multimodal world knowledge in the initial training phase. The pre-training phase typically requires large-scale image-text pair data, such as caption data. For a given image, the model is trained to autoregressively predict the image's caption using a standard cross-entropy loss. A common approach during pre-training is to keep pre-trained modules (such as the visual encoder and LLM) frozen and train the vision-language connector \cite{hu2023rsgpt}. The idea is to align the visual and linguistic modalities without losing the knowledge acquired during pre-training. Some methods \cite{zhan2024skyeyegpt,zhang2024earthgpt,muhtar2024lhrsbot} also unfreeze LLM to allow more trainable parameters for alignment. ShareGPT4V \cite{chen2023sharegpt4v} found that having high-quality caption data during pre-training and unlocking visual encoding capabilities can promote better alignment. In the existing VLGFM landscape, only H$^2$RSVLM \cite{pang2024h2rsvlm} has attempted to unfreeze all modules simultaneously for pre-training.

\noindent$\bullet$
\textbf{Instruction-tuning.} Instruction-tuning aims to teach models to understand user commands better and complete desired tasks. Through this approach, LLMs can be generalized to support unseen tasks, enhancing zero-shot performance. This simple yet effective idea has spurred the success of subsequent NLP works, such as InstructGPT \cite{instructgpt}. Supervised fine-tuning methods typically require much training data to train task-specific models. Instruction-tuning methods reduce the reliance on large-scale data and can accomplish specific tasks through prompt engineering. Due to the flexible format of instructional data and the variability in templates across different tasks, collecting data samples is often more challenging and costly. Recent studies have shown that the quality of instructional fine-tuning samples is more critical than quantity. \cite{zeng2023matters} discovered that high-quality instruction-tuning data can yield better performance, even if the quantity of samples is not the largest.

\subsubsection{Generative VLGFMs}

A typical generative VLGFM comprises three main components: a text encoder, a pre-trained conditional diffusion model, and an image decoder. A schematic of this architecture is depicted in Figure \ref{fig:arch}. The working mechanism of diffusion models consists of two phases. Initially, noise is introduced into the dataset, which is a crucial step in the forward diffusion process, followed by a systematic reversal of this process. In other words, the method involves identifying a stochastic differential equation (SDE) \cite{song2021scorebased} whose process converges to a standard normal distribution. This equation transforms the data, mapping it onto the standard normal distribution. Subsequently, samples that conform to the original data distribution are generated by sampling from this distribution and applying the inverse of the SDE to the sampled data.

Conditional diffusion models are a class of generative models that extend the fundamental principles of diffusion processes to incorporate conditional information, thereby enabling the generation of data that adheres to specific conditions or constraints. In the field of remote sensing, \revii{DiffusionSat}~\cite{khanna2023diffusionsat} successfully enabled the model to generate satellite images of specific locations and seasons by incorporating information such as latitude, longitude, and the time of capture during the training process of the diffusion model. Furthermore, CRS-Diff \cite{tang2024crsdiff} attempted to guide the diffusion model in generating RSIs with similar layouts by using binary masks or road sketches, which can be used to generate samples that are difficult to collect in the real world. Due to limitations in the number of training samples and computational resources, the work on generative models in the domain of remote sensing is still relatively scarce.

\begin{figure}
\centering
\includegraphics[width=9cm]{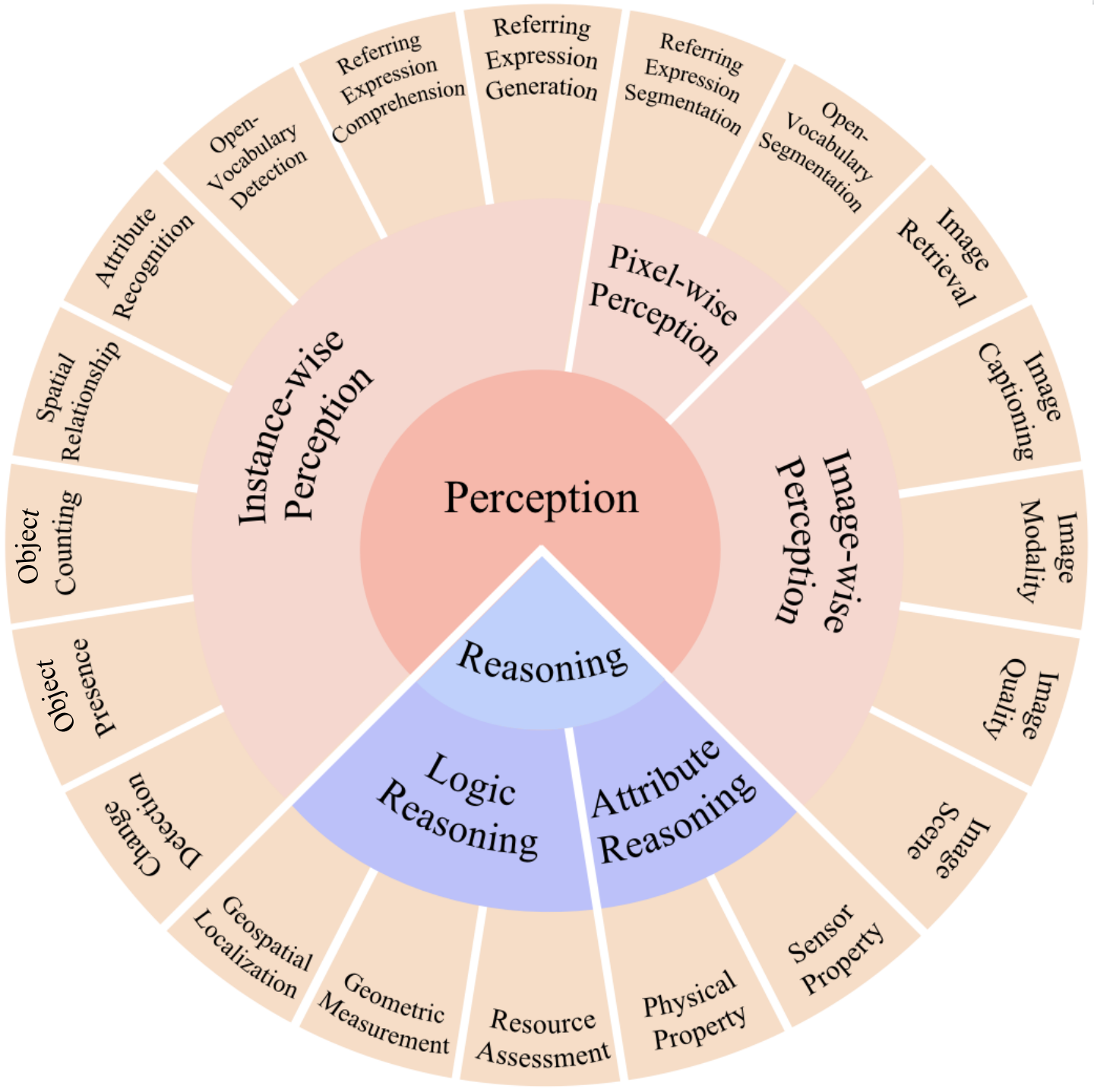}
\caption{\revii{Overview of hierarchical VLGFM capability levels from $L_0$ to $L_2$.}}
\label{fig:capabilities}
\end{figure}

\begin{figure*}
\centering
\includegraphics[width=15cm]{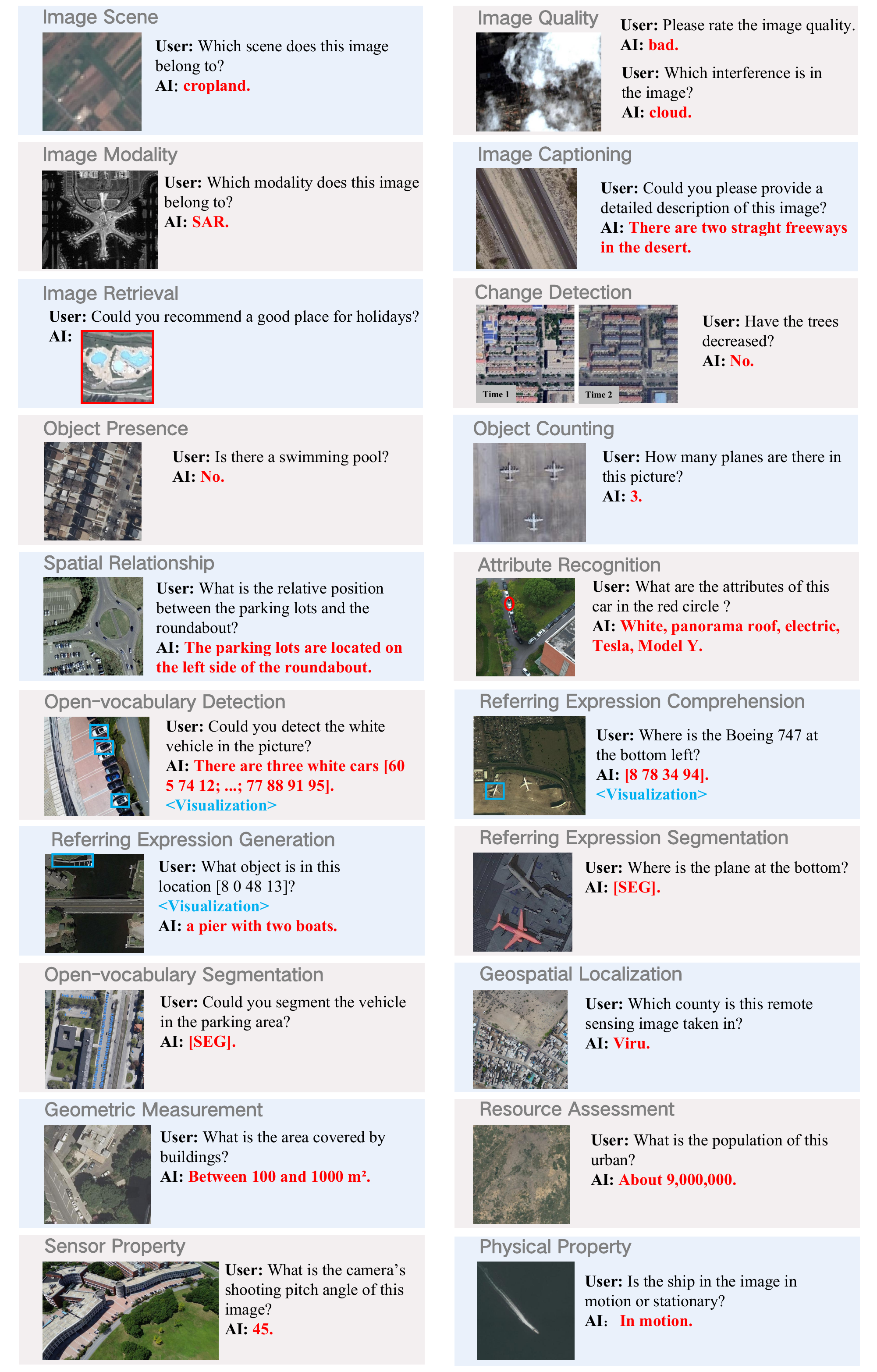}
\caption{\revii{Examples of VLGFM capabilities.}}
\label{fig:demo}
\end{figure*}

\begin{table*}[!h]
\centering 
\vskip 1mm
\caption{\revii{Capabilities supported by major remote sensing vision-language benchmarks. The fourteen displayed acronyms are defined in column order: IS, image scene; IC, image captioning; IR, image retrieval; CD, change detection; OP, object presence; OC, object counting; SR, spatial relationship; AR, attribute recognition; REC, referring expression comprehension; REG, referring expression generation; RES, referring expression segmentation; OVS, open-vocabulary segmentation; GM, geometric measurement; and PP, physical property.}}
\vskip 1mm
\resizebox{\linewidth}{!}{
\begin{tabular}{c|ccclcccccccccc|c}
\hline
Dataset & IS & IC & IR & CD & OP & OC & SR & AR & REC & REG & RES & OVS & GM & PP & Total \\ \hline
Sydney-Captions \cite{sydneycaption} &  & \checkmark &  &  &  &  &  &  &  &  &  &  &  &  & 1 \\
UCM-Captions \cite{sydneycaption} &  & \checkmark & \checkmark &  &  &  &  &  &  &  &  &  &  &  & 2 \\
RSICD \cite{rsicd}&  & \checkmark & \checkmark &  &  &  &  &  &  &  &  &  &  &  & 2 \\
RSVQA-LR \cite{rsvqa} & \checkmark &  &  &  & \checkmark & \checkmark & \checkmark & \checkmark &  &  &  &  &  &  & 5 \\
RSVQA-HR \cite{rsvqa} & \checkmark &  &  &  & \checkmark & \checkmark & \checkmark & \checkmark &  &  &  &  & \checkmark &  & 6 \\
RSVQAxBEN \cite{rsvqaben}& \checkmark &  &  &  & \checkmark &  &  &  &  &  &  &  &  &  & 2 \\
FloodNet \cite{9460988}&  &  &  &  &  & \checkmark &  &  &  &  &  &  &  & \checkmark & 2 \\
RSITMD \cite{rsitmd}&  & \checkmark & \checkmark &  &  &  &  &  &  &  &  &  &  &  & 2 \\
RSIVQA \cite{rsivqa}& \checkmark &  &  &  & \checkmark & \checkmark &  & \checkmark &  &  &  &  &  &  & 4 \\
NWPU-Captions \cite{nwpucaption}&  & \checkmark &  &  &  &  &  &  &  &  &  &  &  &  & 1 \\
LEVIR-CC \cite{levir-cc}& & \checkmark &  & \checkmark &  & &  &  &  &  &  &  & &  & 2 \\
CDVQA \cite{cdvqa}& \checkmark &  &  & \checkmark &  & \checkmark &  &  &  &  &  &  & \checkmark &  & 4 \\
UAV \cite{uavcaption}&  & \checkmark &  &  &  &  &  &  &  &  &  &  &  &  & 1 \\
RSVG \cite{rsvg}&  &  &  &  &  &  & \checkmark & \checkmark & \checkmark & \checkmark &  &  & \checkmark &  & 5 \\
CRSVQA \cite{crsvqa}& \checkmark &  &  &  & \checkmark & \checkmark & \checkmark & \checkmark &  &  &  &  &  &  & 5 \\
DIOR-RSVG \cite{dior-rsvg}&  &  &  &  &  &  & \checkmark & \checkmark & \checkmark & \checkmark &  &  &  &  & 4 \\
RSIEval \cite{hu2023rsgpt} & \checkmark &  &  &  & \checkmark & \checkmark & \checkmark & \checkmark &  &  &  &  &  & \checkmark & 6 \\
SATIN \cite{roberts2023satin}& \checkmark &  &  &  &  &  &  &  &  &  &  &  &  &  & 1 \\
NAIP-OSM \cite{mall2023remote}& \checkmark &  & \checkmark &  &  &  &  &  &  &  &  &  &  &  & 2 \\
RemoteCount \cite{liu2024remoteclip} &  &  &  &  &  & \checkmark &  &  &  &  &  &  &  &  & 1 \\
SkyScript \cite{wang2024skyscript}& \checkmark &  & \checkmark &  &  &  &  &  &  &  &  &  &  &  & 2 \\
EarthVQA \cite{wang2024earthvqa}& \checkmark &  & \checkmark &  &  & \checkmark & \checkmark &  &  &  &  &  &  & \checkmark & 5 \\
RRSIS \cite{rrsis}&  &  &  &  &  &  &  &  &  &  &  & \checkmark &  &  & 1 \\
GeoChat-Bench \cite{kuckreja2023geochat}& \checkmark &  &  &  &  & \checkmark & \checkmark & \checkmark & \checkmark & \checkmark &  &  &  &  & 6 \\
RRSIS-D \cite{rrsis-d}&  &  &  &  &  &  &  &  &  &  & \checkmark &  &  &  & 1 \\
ChatEarthNet \cite{yuan2024chatearthnet}& \checkmark & \checkmark &  &  &  &  & \checkmark &  &  &  &  &  &  &  & 3 \\ 
VLEO-Bench \cite{vleobench}& \checkmark & \checkmark &  & \checkmark &  & \checkmark &  &  & \checkmark & \checkmark &  &  &  &  & 6 \\
FineGrip \cite{finegrip}&  & \checkmark &  &  &  &  &  &  &  &  &  & \checkmark &  &  & 2 \\
RS-GPT4V \cite{xu2024rsgpt4v}&  & \checkmark &  &  & \checkmark &  & \checkmark & \checkmark & \checkmark & \checkmark &  &  &  &  & 6 \\
VRSBench \cite{li2024vrsbench}&  \checkmark & \checkmark &  &  & \checkmark & \checkmark &  & \checkmark & \checkmark & \checkmark &  &  &  &  & \textbf{7} \\ \hline
\end{tabular}}
\label{Table-benchmark}
\end{table*}

\subsection{Capabilities}
\label{sec:capabilities}

Inspired by MMBench \cite{liu2023mmbench}, we have categorized the capabilities exhibited by VLGFMs into three hierarchical levels and clarified the design principles underlying this classification.

\noindent \textbf{Overview.} Figure \ref{fig:capabilities} provides an overview of the existing multimodal capabilities of VLGFM. This benchmark encompasses 20 different leaf capabilities, offering a comprehensive evaluation of various fine-grained perceptual and reasoning abilities. \revii{Figure~\ref{fig:demo} illustrates these capabilities, while Table~\ref{Table-benchmark} identifies the capabilities covered by major remote sensing vision-language benchmarks. Capabilities without a dedicated evaluation resource are retained in the taxonomy and explicitly marked in the complete mapping below.} \hl{The taxonomy is intended to be extensible and task-oriented rather than strictly closed or mutually exclusive. A capability is assigned to a leaf according to the primary evidence required by the model output: direct visual evidence for perception, inferred or external knowledge for reasoning, and pixel-level output for segmentation-related perception. Borderline cases are explicitly acknowledged because many geospatial tasks combine visual perception with geographic reasoning.}

\noindent \textbf{The Hierarchical Taxonomy of Capability.} Humans possess remarkable perceptual and reasoning abilities, enabling them to understand and interact with the world. These capabilities have been crucial in human evolution and are the foundation for complex cognitive processes. Perception involves gathering information from sensory inputs, while reasoning involves drawing conclusions based on this information. Collectively, they underpin the foundation of most tasks in the real world, including object recognition, problem-solving, and decision-making \cite{fodor1983modularity,oaksford2007bayesian}. VLGFMs are also expected to exhibit strong perceptual and reasoning capabilities in pursuit of artificial intelligence for remote sensing verticals. Therefore, our capability taxonomy designates perception and reasoning as our top-level capability dimensions, referred to as L-1 capability dimensions. For L-2 capabilities, we derive image-wise perception, region-wise perception, instance-wise perception, pixel-wise perception from L-1 perception, Attribute reasoning, and logical reasoning from L-1 reasoning. To make our benchmark as granular as possible, we further derive 20 L-3 capability dimensions from L-2. \revii{Table~\ref{tab:capability-mapping} contains one row for every L-3 leaf and maps it to representative models, evaluation resources, and the evidence or output used for assignment.}

\begin{table*}[!t]
\centering
\scriptsize
\caption{Mapping of the twenty L-3 capability leaves to representative VLGFMs, evaluation resources, and the evidence required for assignment. A dash indicates that no dedicated benchmark was identified.}
\label{tab:capability-mapping}
\setlength{\tabcolsep}{2.5pt}
\renewcommand{\arraystretch}{1.03}
\begin{tabularx}{\linewidth}{>{\RaggedRight\arraybackslash}p{2.0cm}>{\RaggedRight\arraybackslash}p{2.8cm}>{\RaggedRight\arraybackslash}p{3.7cm}>{\RaggedRight\arraybackslash}p{4.1cm}>{\RaggedRight\arraybackslash}X}
\toprule
\textbf{L-2 group} & \textbf{L-3 capability} & \textbf{Representative model(s)} & \textbf{Evaluation resource(s)} & \textbf{Assignment evidence} \\
\midrule
\multirow{5}{=}{Image-wise} & Image scene & RemoteCLIP, GeoRSCLIP, SkyScript & AID, EuroSAT, RESISC45, SATIN & Image-level scene or land-use/cover label \\
& Image quality & Conversational VLGFMs & -- & Cloud, noise, visibility, or quality judgment \\
& Image modality & H$^2$RSVLM & LHRS-Bench/model-specific questions & Optical, SAR, hyperspectral, or other modality label \\
& Image captioning & RSGPT, SkyEyeGPT, RS-CapRet, GeoZero & RSICD, UCM, Sydney, NWPU & Free-form whole-image description \\
& Image retrieval & RemoteCLIP, GeoRSCLIP, RS-CapRet & RSICD, RSITMD, UCM & Ranked image--text matches evaluated by R@K \\
\midrule
\multirow{8}{=}{Instance-wise} & Change detection & VLEO-Bench-tested VLGFMs & LEVIR-CC, CDVQA, VLEO-Bench & Changed object, region, or state \\
& Object counting & RemoteCLIP, H$^2$RSVLM & RemoteCount, RSVQA, VLEO-Bench & Integer count for a named class \\
& Object presence & RSGPT, GeoChat, RS-GPT4V & RSVQA, RSIVQA, VRSBench & Binary target-presence decision \\
& Spatial relationship & GeoChat, SkySenseGPT & GeoChat-Bench, RSVQA, CRSVQA & Visually observable relative or absolute relation \\
& Attribute recognition & GeoChat, SkyEyeGPT, RSCoVLM & GeoChat-Bench, DIOR-RSVG, VRSBench & Visible color, size, shape, material, or orientation \\
& Open-vocabulary detection & SkySenseGPT, EarthDial, RingMo-Agent, RSCoVLM & VRSBench/model-specific sets & Category- or text-conditioned bounding boxes \\
& Referring expression comprehension & GeoChat, SkyEyeGPT, LHRS-Bot, GeoGround & RSVG, DIOR-RSVG, VRSBench, AVVG & Box or coordinates for one referred object \\
& Referring expression generation & GeoChat, SkyEyeGPT, SkySenseGPT & GeoChat-Bench, RSVG, DIOR-RSVG & Discriminative description of a selected region \\
\midrule
\multirow{2}{=}{Pixel-wise} & Referring expression segmentation & H$^2$RSVLM, segmentation VLGFMs & RRSIS-D/model-specific sets & Text-conditioned binary mask \\
& Open-vocabulary segmentation & FineGrip, segmentation VLGFMs & RRSIS, FineGrip & Category masks beyond a closed vocabulary \\
\midrule
\multirow{3}{=}{Logical reasoning} & Geospatial localization & RS-GPT4V, general VLGFMs & RS-GPT4V/GPT4Geo-style tasks & Place name or coordinates supported by geographic cues \\
& Geometric measurement & H$^2$RSVLM, AirSpatialBot, GPT-4V studies & AVI-Math, AirSpatial SQA/3DBB, GPT4Geo-style tasks & Metric estimate supported by calibrated geometry or 3D reference, with units and error \\
& Resource assessment & VLGFMs with external estimators & Population, crop, or carbon tasks & Quantity or distribution supported by domain priors/data \\
\midrule
\multirow{2}{=}{Attribute reasoning} & Sensor property & H$^2$RSVLM & Model-specific questions & Metadata or qualified estimate of time, GSD, view, or focal length \\
& Physical property & H$^2$RSVLM, conversational VLGFMs & FloodNet, RSIEval, EarthVQA & Physical state inferred from visual evidence and domain knowledge \\
\bottomrule
\end{tabularx}
\end{table*}

\noindent \textbf{Taxonomy boundary.}
\hl{Spatial relationship is placed under perception in this version because many current benchmarks ask for visually observable left-right, near-far, or inside-outside relations. However, spatial relationship understanding becomes a reasoning capability when it involves map priors, metric distance, sensor geometry, or multi-view consistency. We therefore treat the taxonomy as an analytical tool for organizing evidence rather than a rigid ontology.}

\subsubsection{Perception}

Perceptual capability is the foundational skill for interpreting RSIs. With the rise of deep learning technology, numerous specialized models have been proposed to address image perception tasks in remote sensing, such as scene classification \cite{roberts2023satin}, object detection \cite{mmrotate}, and segmentation \cite{rsi-seg}. The emergence of VLGFM provides a solution for handling all perception tasks using a single model. Perception can be mainly divided into three types: image-wise, instance-wise, and pixel-wise perception.

\revii{Image-wise perception belongs to coarse-grained perception and is relatively easier.} It can be divided into image scene, image quality, image modality, image captioning, and image retrieval. Instance-wise perception belongs to fine-grained perception, which requires identifying objects within an image and is more challenging. It can be divided into eight categories, encompassing some visual grounding capabilities. The basic building blocks of an image are pixels; thus, pixel-wise perception represents the most fine-grained perceptual capability. Pixel-wise perception often requires an additional pixel decoder. The SAM decoder \cite{SAM} is a common choice. The output features of VLGFM will be aligned in dimensions and then concatenated with the prompt features of SAM. Finally, they are fed into the SAM decoder to generate the mask. Pixel-wise perception can be divided into referring expression segmentation and open-vocabulary segmentation.

\noindent$\bullet$ \textbf{Image Scene}. \revii{Scene recognition categorizes an image into a predefined set of scenes or environments based on its overall visual content.} This review, referencing the SATIN framework \cite{roberts2023satin}, categorizes remote sensing scenes into six distinct categories. Land cover refers to the biophysical surface characteristics of the Earth, with examples including forests, grasslands, and buildings. Land use describes the economic and social functions of areas, such as residential zones, industrial sectors, and parking lots, including bodies of water utilized for specific purposes. Hierarchical land use takes this a step further by classifying land use into more detailed levels of granularity, for instance, categorizing an area broadly as a transportation area, then more specifically as highway land, and even more precisely as a bridge. Complex scenes are images that contain one or more land use labels, indicating a blend of different uses or covers in one scene. Rare scenes represent unusual or infrequent categories that do not typically fall under standard land cover or land use definitions, such as hurricane damage or wind turbines. Lastly, false color scenes are images represented in false color, focusing on a single land use or land cover class, aiding in the distinct visualization of specific area characteristics.

\noindent$\bullet$ \textbf{Image Quality}. \revii{Image-quality assessment evaluates RSI conditions such as imaging quality, noise \cite{alamri2010noise}, cloud cover \cite{SHEN2014cloud}, and low-light conditions \cite{lowlight}.} This capability helps users filter out low-quality RSIs to prevent noise from contaminating the results. To date, no remote sensing benchmarks are specifically designed to test image quality capabilities.

\noindent$\bullet$ \textbf{Image Modality}. \revii{Image-modality recognition classifies remote sensing imagery as optical, hyperspectral, synthetic aperture radar, or another modality.} This capability enables agent models to identify the modality of an image, thereby invoking the appropriate interpretation tools for that modality. Currently, no remote sensing benchmarks are specifically designed to test image modality capabilities.

\noindent$\bullet$ \textbf{Image Captioning}. \revii{Image captioning produces a coherent and contextually relevant narrative in natural language that encapsulates the overall scene depicted in an image.} Image captioning can further be categorized based on the input type into change captioning \cite{10271701}, multi-temporal image captioning, and video captioning \cite{capera}. The input for the change captioning task consists of two images. The objective is to summarize and describe the changes observed in the later image relative to the earlier one in the temporal sequence. Compared to change captioning, which involves two images, multi-temporal image captioning deals with a larger set of images and aims to encapsulate the progression of changes exhibited across a series of temporal images. In the case of video captioning, the input is a video sequence, and the task is to describe the events occurring within the video.

\noindent$\bullet$ \textbf{Image Retrieval}. \revii{Image retrieval finds images from a large remote sensing database that are relevant to a given query image or text.} Since this paper focuses on vision-language models, it considers image retrieval using text queries. In practical applications, users can quickly pinpoint specific RSIs through textual descriptions.

\noindent$\bullet$ \textbf{Change Detection}. \revii{Change detection identifies areas or objects that have changed in RSIs of the same place at different times.} Compared to change captioning capabilities, it focuses more on specific instances in the image than the entire image. Observing changes in specific targets is a more practical function in real-world applications. LEVIR-CC \cite{levir-cc} and CDVQA \cite{cdvqa} analyze changes in vegetation and buildings, which can be used to track the progression of urbanization.

\noindent$\bullet$ \textbf{Object Counting}. \revii{Object counting identifies and counts instances of specified objects within an image.} Due to the high density of targets typically present in RSIs, manual techniques are time-consuming and labour-intensive. This makes counting the number of targets in RSIs a particularly valuable capability. For example, we can count animals, trees, buildings, and vehicles in RSIs \cite{vleobench}.

\noindent$\bullet$ \textbf{Object Presence}. \revii{Object-presence recognition determines whether a specific target type exists within an image.} Essentially, this capability requires the AI to answer whether the number of such targets is zero, which overlaps somewhat with object counting. However, detecting the presence of targets is a common requirement in real-world applications, which is why it is recognized as a distinct leaf capability.

\noindent$\bullet$ \textbf{Spatial Relationship}. \revii{Spatial-relationship recognition describes the absolute positional relationships of targets relative to the image or their relative positional relationships to other targets within the image.} In general vision-language foundation models, this capability is often considered a form of reasoning. However, because the positional relationships of targets in RSIs are relatively simple, we categorize it as a perceptual capability.

\noindent$\bullet$ \textbf{Attribute Recognition}. \revii{Attribute recognition identifies the attributes of a specific target within an image, including aspects of its appearance such as colour, size, and material.} For instance, detailed attribute features of a car can help determine the vehicle's brand and even its specific model. This capability is highly dependent on the resolution of RSIs; higher resolution allows for capturing more detailed attributes of the target.

\noindent$\bullet$ \textbf{Open-vocabulary Detection}. \revii{Open-vocabulary detection detects and recognizes objects, including those not encountered during training, by leveraging a flexible and expansive vocabulary.} It can be considered an advanced version of traditional object detection. It detects common object categories in remote sensing datasets and identifies objects with descriptive constraints. For instance, one can detect all vehicles in an image or specifically all white vehicles. The detection results are returned in textual format.

\noindent$\bullet$ \textbf{Referring Expression Comprehension}. \revii{Referring expression comprehension localizes a target object in an image described by a natural-language expression.} The difference from open-vocabulary detection is that referring expression comprehension refers to a single target, whereas open-vocabulary detection can output multiple targets of the same type. The detection results are returned in the same text format as the training data.

\noindent$\bullet$ \textbf{Referring Expression Generation}. \revii{Referring expression generation produces a discriminative description of an object in an image.} It is very similar to image captioning. The distinction from attribute recognition is that attribute recognition describes the features of a single target, whereas referring expression generation describes the content within a specified bounding box.

\noindent$\bullet$ \textbf{Referring Expression Segmentation}. \revii{Referring expression segmentation segments a target object in an image described by a natural-language referring expression.} It can be considered the semantic segmentation version of referring expression comprehension. Unlike referring expression comprehension, its output is a binary mask image.

\noindent$\bullet$ \textbf{Open-vocabulary Segmentation}. \revii{Open-vocabulary segmentation segments and classifies image regions using a comprehensive and extendable set of labels not limited by a predefined vocabulary.} Unlike open-vocabulary detection, its output is a binary mask image. RSIs encompass visible-light, multispectral, and hyperspectral data. Effectively utilizing this information for segmentation presents a significant challenge.

\subsubsection{Reasoning}

\revii{Perception tasks can be accomplished using specialized vision models, while autonomous reasoning is the capability most eagerly anticipated in remote sensing interpretation~\cite{roberts2024charting}.} Reasoning poses greater difficulty and has a higher application value than perception. Reasoning can be mainly divided into two types: logical and attribute reasoning.

Logical reasoning requires leveraging expert knowledge from specific domains to complete the inference process. This external information can include geographical, cartographic, anthropological, and agricultural knowledge. Logical reasoning capabilities fully demonstrate the advantages of VLGFMs and can be divided into geospatial localization, geometric measurement, and resource assessment. Compared to attribute recognition in perceptual capabilities, the attributes in attribute reasoning cannot be directly observed from the image but require inference combined with sensor parameters and physical knowledge. Attribute reasoning can be divided into sensor property and physical property.

\noindent$\bullet$ \textbf{Geospatial Localization}. \revii{Geospatial localization pinpoints the geographic location represented in an RSI, potentially identifying the continent, country, city, or precise latitude and longitude based on terrain and landmark features~\cite{NEURIPS2023_1b57aadd}.} Its inspiration comes from the popular geography game GeoGuessr\footnote{https://www.geoguessr.com}, in which players are placed in a street-view panorama somewhere in the world and must guess their location. Geospatial localization typically requires acquiring information on topography, vegetation, and building styles through perceptual capabilities to perform reasoning, which adds significant complexity.

\noindent$\bullet$ \textbf{Geometric Measurement}. \revii{This capability estimates distance, length, area, height, angle, or elevation from remote sensing observations~\cite{roberts2023gpt4geo}. A metric result is defensible only when an image-to-ground transformation or equivalent scale and geometry are available. For a map-projected orthorectified image with ground pixel sizes $g_x$ and $g_y$, a straight segment lying on the reference surface may be measured as $L=\sqrt{(g_x\Delta u)^2+(g_y\Delta v)^2}$, and a region containing $N_p$ pixels may be approximated as $A=N_p g_xg_y$. For square pixels, these expressions reduce to multiplication by the GSD and GSD$^2$, respectively. They provide planimetric measurements on the orthoimage reference surface, not three-dimensional surface distance or object height, and their uncertainty also depends on boundary localization, georegistration, terrain representation, and residual structure displacement. Shadow-based height, $h\approx L_s\tan\alpha$, additionally requires a ground-referenced shadow length $L_s$, known solar elevation $\alpha$, a locally level surface, a vertical object, and an unambiguous shadow endpoint.}

\revii{Two complementary routes are emerging for aerial geometric reasoning. The first uses explicit imaging geometry. AVI-Math supplies camera intrinsics, sensor pixel size, above-ground-level altitude, and pitch angle, and transforms detected pixel locations into camera or ground coordinates under a flat-ground assumption before computing distance, area, orientation, or trajectory quantities~\cite{zhou2025avimath}. This design makes the reasoning chain auditable, but it also separates two failure sources: visual localization of the relevant objects and numerical/geometric computation from the supplied parameters. Its assumptions must be checked for terrain relief, camera motion, inaccurate altitude or attitude, and non-ground objects.}

\revii{The second route learns spatial quantities from 3D supervision. AirSpatialBot introduces spatial question answering for depth, distance, length, width, and height, together with spatial grounding based on three-dimensional bounding boxes~\cite{zhou2025airspatialbot}. Its two-stage training transfers 2D localization knowledge to 3D prediction and imposes consistency between predicted 2D and 3D coordinates; an agent then combines spatial estimates with task planning and external attribute tables. Such learned estimates are useful when an explicit photogrammetric solution is unavailable, but they remain dependent on the training distribution and reference-coordinate construction. Their geometric validity should therefore be assessed using coordinate or dimension errors and 3D/BEV overlap, rather than inferred only from downstream brand recognition or retrieval accuracy.}

\revii{More generally, raw or oblique imagery does not have a uniform ground scale. A single uncalibrated view is insufficient to recover unique metric three-dimensional geometry. Classical reconstruction requires an appropriate sensor/camera model, orientation, lens-distortion treatment, terrain or surface information, and a source of absolute scale; bundle adjustment additionally requires overlapping images and obtains metric scale only through control points, a known baseline, GNSS/IMU observations, or another constraint. A measurement-oriented VLGFM should select between explicit geometric calculation, learned 3D estimation, and external GIS or photogrammetric tools according to the available evidence, rather than generate a number from appearance alone.}

\revii{The output should report the coordinate reference system and datum, units, product level, scale or GSD source, sensor model, reference surface, measurement procedure, and assumptions. Spatial resolution must not be presented as positional accuracy. Accuracy should be evaluated using independent checkpoints or reference measurements not used in model fitting, together with their accuracy and spatial distribution. For point coordinates, appropriate statistics include mean signed errors and RMSE$_x$, RMSE$_y$, and RMSE$_z$; horizontal and three-dimensional summaries can be computed as $\mathrm{RMSE}_H=\sqrt{\mathrm{RMSE}_x^2+\mathrm{RMSE}_y^2}$ and $\mathrm{RMSE}_{3D}=\sqrt{\mathrm{RMSE}_x^2+\mathrm{RMSE}_y^2+\mathrm{RMSE}_z^2}$ when all components are available. The ASPRS standard emphasizes declared, fit-for-purpose accuracy classes and the contribution of checkpoint/reference uncertainty rather than a universal tolerance~\cite{asprs2024accuracy}. An operational VLGFM should propagate uncertainty from image localization, scale, sensor geometry, control, and reference data; refuse unsupported metric requests; and distinguish semantic localization from ground-space positional accuracy.}

\noindent$\bullet$ \textbf{Resource Assessment}. \revii{Resource assessment estimates and analyzes the quantity, quality, and spatial distribution of natural resources.} Examples include estimating urban population sizes~\cite{roberts2024charting}, calculating forest carbon dioxide emissions, and estimating crop yields.

\noindent$\bullet$ \textbf{Sensor Property}. \revii{Sensor-property reasoning retrieves or estimates acquisition time, platform altitude and direction, GSD, focal length, view angle, spectral modality, and related metadata. Parameters inferred from pixels alone should be labeled as estimates because different camera configurations can produce similar projections. Whenever metadata are available, the model should cite them and propagate their uncertainty into oblique measurement or monocular 3D reconstruction rather than treating inferred values as calibration.}

\noindent$\bullet$ \textbf{Physical Property}. \revii{Physical-property reasoning infers the physical characteristics or conditions of objects within an image, such as moving targets or houses submerged by a flood.} These properties are often not directly obtainable from the image and require inference combined with physical common sense. For example, rust on the roof of a building may indicate that the roof is made of iron, while a sailing ship leaves a wake behind it.

\section{Challenges and Outlook}

\subsection{Challenges}

\noindent\textbf{Overview.}
\hl{The main challenges of VLGFMs are not only inherited from general multimodal models, but also rooted in the imaging mechanisms, spatial scales, sensor diversity, and measurement requirements of earth observation. We therefore reorganize the challenges around remote-sensing-specific failure modes.}

\noindent$\bullet$
\textbf{Small, dense, and multi-scale object understanding.} \hl{Remote sensing images often contain dense vehicles, ships, buildings, trees, and small infrastructure objects. These targets may occupy only a few pixels, while their semantic importance can be high. Cropping or down-sampling high-resolution RSIs may remove fine details, whereas processing full-resolution scenes is computationally expensive. This tension directly affects captioning specificity, object counting, referring expression comprehension, and open-vocabulary detection.}

\noindent$\bullet$
\textbf{Noise and bias in geo-language supervision.} \hl{Large-scale VLGFM datasets are frequently constructed from templates, OSM tags, social-media captions, or LLM-generated question-answer pairs. These sources improve scalability, but they may introduce generic captions, hallucinated object relations, inaccurate spatial expressions, and geographic or class imbalance. As a result, a model may learn dataset-specific linguistic shortcuts rather than robust geospatial semantics.}

\noindent$\bullet$
\textbf{Limited coverage of non-RGB and temporal modalities.} \hl{Earth observation includes optical, infrared, SAR, multispectral, hyperspectral, LiDAR, DEM, and time-series observations. Most current VLGFMs are still centered on RGB-like optical images, leaving important gaps in all-weather sensing, spectral material analysis, elevation-aware reasoning, and temporal change understanding. Effective multimodal fusion remains difficult because different sensors have different resolutions, noise distributions, imaging geometries, and physical meanings.}

\noindent$\bullet$
\textbf{Benchmark insufficiency for geospatial reasoning and measurement.} \hl{Existing benchmarks cover many basic VQA, captioning, retrieval, and grounding tasks, but they still provide limited evidence for metric reliability, sensor-parameter inference, multi-view consistency, and physically grounded reasoning. High scores on simple question-answer tasks may not indicate readiness for mapping, monitoring, emergency response, or other operational applications.}

\noindent$\bullet$
\textbf{Hallucination, uncertainty, and reliability.} \hl{Hallucination is particularly risky in remote sensing because generated descriptions may be used for critical target interpretation, disaster assessment, or map updating. A reliable VLGFM should not only answer questions, but also estimate uncertainty, refuse under-specified queries, cite visual evidence, and remain consistent with geospatial constraints.}

\noindent$\bullet$
\textbf{Geometric and photogrammetric reliability.} \hl{For photogrammetric and measurement-oriented applications, semantic correctness alone is insufficient. VLGFMs must reason about scale, orientation, camera geometry, ground sampling distance, oblique viewing, shadows, occlusion, and sensor metadata. Current models can describe objects and rough spatial relations, but they are not yet dependable tools for precise geometric measurement, monocular 3D reconstruction, or sensor-parameter inference without explicit geometric constraints.}

\subsection{Future Work}

\noindent\textbf{Overview.}
\hl{Future work should be derived from the empirical and conceptual gaps identified above rather than from generic trends in large multimodal models. We rank the following directions according to their relevance to practical earth observation and their ability to close current capability gaps.}

\noindent$\bullet$
\textbf{Priority 1: reliable grounding and small-object understanding.} \hl{Future VLGFMs should improve object-level localization, dense target counting, and referring expression comprehension in high-resolution scenes. This requires better coordinate representations, multi-scale visual encoders, resolution-preserving training, and benchmarks that evaluate both semantic correctness and localization accuracy.}

\noindent$\bullet$
\textbf{Priority 2: multimodal and multi-sensor VLGFMs.} \hl{A central direction is to move beyond RGB optical imagery toward unified modeling of optical, infrared, SAR, multispectral, hyperspectral, LiDAR, DEM, and time-series data. Such models should align heterogeneous modalities while preserving their physical meaning, uncertainty, and spatial resolution.}

\noindent$\bullet$
\textbf{Priority 3: geometry-aware and measurement-reliable VLGFMs.} \hl{For photogrammetry, mapping, and monitoring, VLGFMs should incorporate sensor metadata, ground sampling distance, camera orientation, and geometric constraints. Promising directions include tool-augmented measurement, uncertainty-aware metric reasoning, oblique-view understanding, and integration with 3D reconstruction pipelines.}

\noindent$\bullet$
\textbf{Priority 4: remote sensing AI agents.} \hl{Remote sensing agents can combine VLGFMs with GIS tools, geospatial databases, retrieval systems, object detectors, segmentation models, change detection modules, and external reasoning tools. Such agents may support multi-step tasks such as disaster monitoring, urban analysis, automatic report generation, and interactive map updating. However, their evaluation must consider tool-use reliability, evidence grounding, safety, and reproducibility.}

\noindent$\bullet$
\textbf{Priority 5: trustworthy data construction and evaluation.} \hl{The field needs datasets with transparent provenance, balanced geographic coverage, high-quality annotations, and explicit uncertainty labels. Evaluation should move from isolated accuracy scores toward capability-specific protocols that measure robustness, calibration, refusal behavior, hallucination rate, and cross-region generalization.}

\noindent$\bullet$
\textbf{Priority 6: efficient adaptation and training-free methods.} \hl{Because training full VLGFMs remains expensive, parameter-efficient tuning, retrieval-augmented generation, prompt-based reasoning, and training-free tool use will remain important. These methods should be evaluated not only by average performance, but also by whether they preserve geospatial consistency and reduce hallucination under domain shift.}

\section{Summary and Conclusions}

This survey provides a detailed examination of the latest developments in VLGFMs. We outline the necessary background knowledge, including the fundamental concepts and introductory information about VLGFM. Subsequently, we summarize two data collection methods, three model architectures, and twenty fundamental capabilities. For each capability, we provide straightforward descriptions and intuitive examples. \hl{In the revised organization, we further position this survey against related reviews, clarify the literature collection protocol, add quantitative synthesis tables, discuss the trade-offs of data construction strategies, and emphasize remote-sensing-specific challenges such as small-object grounding, multimodal fusion, hallucination, and geometric reliability.} Finally, we summarize several challenges and highlight some future research directions for VLGFMs.


%

	
\ifCLASSOPTIONcaptionsoff
\newpage
\fi

\bibliographystyle{IEEEtran}
\bibliography{main}


\end{document}